\documentclass[twoside]{article}

\usepackage[accepted]{aistats2026}

\usepackage[utf8]{inputenc} %
\usepackage[T1]{fontenc}    %
\usepackage{hyperref}       %
\usepackage{url}            %
\usepackage{booktabs}       %
\usepackage{amsfonts}       %
\usepackage{nicefrac}       %
\usepackage{microtype}      %
\usepackage{xcolor}         %
\usepackage{algorithm2e}
\usepackage{amsmath}
\usepackage{graphicx}
\usepackage{wrapfig}
\usepackage{tcolorbox}
\usepackage{subcaption}
\usepackage{soul}         %
\usepackage{xcolor}
\tcbuselibrary{listings}
\usepackage{listings}
\usepackage{booktabs} %
\usepackage{enumitem}
\usepackage{multirow}

\definecolor{changebg}{RGB}{200, 220, 255}  %
\definecolor{negbg}{RGB}{255, 200, 200}     %
\definecolor{posbg}{RGB}{200, 255, 200}     %

\newcommand{\remove}[1]{\sethlcolor{changebg}\hl{#1}}
\newcommand{\possent}[1]{\sethlcolor{posbg}\hl{#1}}   
\newcommand{\negsent}[1]{\sethlcolor{negbg}\hl{#1}}  

\usepackage[round]{natbib}
\begin{document}
\runningauthor{Chen, Ko, Zhang, Cho, Chung, Giuffr\'{e}, Shung, Stadie}

\twocolumn[

\aistatstitle{LAMP: Extracting Local Decision Surfaces From Large Language Models}

\aistatsauthor{
  Ryan Chen\textsuperscript{*,1} \And
  Youngmin Ko\textsuperscript{*,1} \And
  Catherine Cho\textsuperscript{1} \And
  Zeyu Zhang\textsuperscript{1} \AND
  Sunny Chung\textsuperscript{2} \And
  Mauro Giuffr\'{e}\textsuperscript{3} \And
  Dennis L. Shung\textsuperscript{4} \And
  Bradly Stadie\textsuperscript{1}
}

\aistatsaddress{
  \textsuperscript{1}Department of Statistics and Data Science, Northwestern University \\
  \textsuperscript{2}Section of Digestive Diseases, Department of Medicine, Yale School of Medicine \\
  \textsuperscript{3}Department of Biomedical Informatics and Data Science, Yale School of Medicine \\
  \textsuperscript{4}Division of Gastroenterology and Hepatology, Department of Medicine, Mayo Clinic
} ]

{
\let\thefootnote\relax
\footnotetext{\textsuperscript{*}Equal contribution.}
}

\begin{abstract}

We introduce \textbf{LAMP} (\textbf{L}ocal \textbf{A}ttribution \textbf{M}apping \textbf{P}robe), a method that shines light onto a black-box language model's decision surface and studies how reliably a model maps its stated reasons to its reported predictions by approximating a decision surface. LAMP treats the model's own self-reported explanations as a coordinate system and fits a locally linear surrogate that links those weights to the model's output. By doing so, it reveals how much the stated factors steer the model's decisions. We apply LAMP to three tasks: \textit{sentiment analysis}, \textit{controversial-topic detection}, and \textit{safety-prompt auditing}. Across these tasks, LAMP reveals that many language models' locally approximated linear decision landscapes overall agree with human judgments on explanation quality and, on a clinical case‑file data set, align with expert assessments. Since LAMP operates without requiring access to model gradients, logits, or internal activations, it serves as a practical and lightweight framework for auditing proprietary language models, and enabling assessment of whether a model appears to behave consistently with the explanations it provides.

\end{abstract}

\section{INTRODUCTION}

\begin{figure*}[htp]
    \centering
    \includegraphics[width=0.9\linewidth]{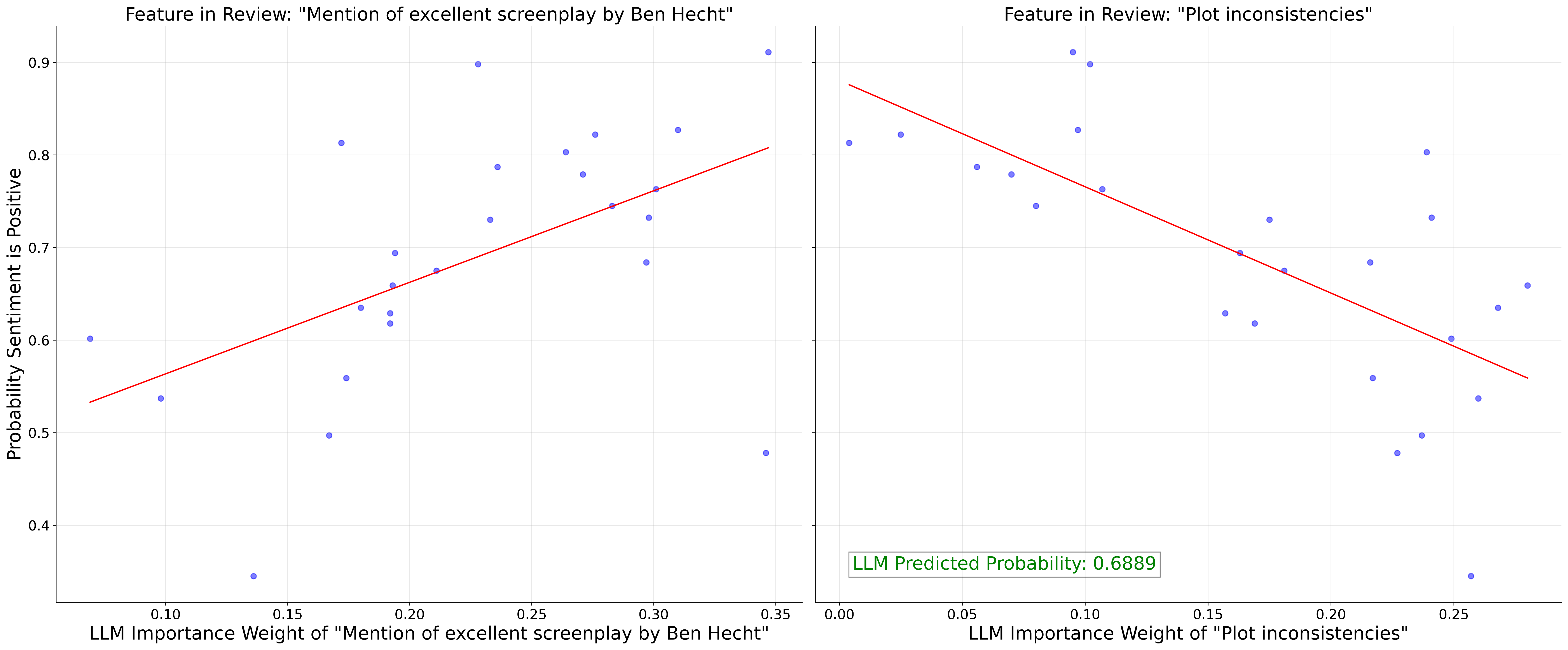}
    \caption{\textbf{LAMP produces features that can model the language model decision surface, locally, as a linear combination.} The decision surface is estimated by sampling perturbations around the input and regressing the resulting probabilities reported by the language model. The decision surface can then be expressed as a linear combination of LAMP-generated factors, providing an interpretable summary of the model's local decision behavior. Here, we show two language model extracted factors that the language model reports to contribute to the sentiment classification.}
    \label{fig:sample}
\end{figure*}

Large language models (LLMs) have been used to produce free-text rationales, natural language explanations that articulate step-by-step reasoning behind a prediction \citep{cot, survey-reasoning-llm}. These rationales are appealing since they are interpretable and can provide auxiliary supervision to improve task performance \citep{sc-cot, wang2025wonderful, masterman2024landscape}.  

However, recent work shows that such rationales are often misleading. Chain-of-thought explanations can systematically rationalize biased or incorrect predictions \citep{turpin2023}, and models often perform just as well with prompts that are irrelevant or pathologically misleading \citep{webson2022}. \citep{min2022} further demonstrate that replacing ground-truth labels in in-context learning demonstrations with random labels barely degrades performance, suggesting models rely more on structural cues than semantic labels. 

These results imply that while rationales can be useful, they do not guarantee that explanations reflect a model's decision process \citep{faith-hard, faith-or-sc}. As the evaluation of textual rationales remains an active research direction, we pivot towards a behavior-centric approach. Rather than adjudicating faithfulness, we characterize the model behavior with a local decision-surface analysis. We study how model behaviors change with small, structured variations in the model-cited rationales, providing an empirical view of consistency.

Rather than evaluating whether model-cited rationales are faithful, we treat these rationales as dimensions on which to evaluate the model's decision landscape, a mapping from self-reported rationales to predicted class probabilities. Our aim is not to validate whether the factors are causally correct, but to study whether the model's predictions vary in a consistent and locally predictable way when those claimed factors are perturbed. By analyzing whether outputs change in structured and predictable patterns, we obtain a geometric picture of model behavior, independent of the model's internal reasoning and activations.

This framework relies on the premise that language models implicitly contain structured internal representations of external world regularities. Recent studies provide compelling evidence supporting this assumption. \citet{guesstimate} shows that models can perform estimation tasks that require approximate knowledge of real-world magnitudes. Additionally, \citet{world-reps} apply state-abstraction theory from reinforcement learning to probe language model world representations. These results motivate the view that language models construct broad, implicit world models, which can be externally probed and characterized.

To this end, we introduce the \textbf{L}ocal \textbf{A}ttribution \textbf{M}apping \textbf{P}robe (\textbf{LAMP}), a framework to extract and analyze local linear approximations of a black-box language model's decision landscape. Specifically, LAMP (i) elicits a weighted set of explanatory factors from the model, (ii) perturbs those weights stochastically and re-queries the model to measure changes in predictions, and (iii) fits a linear surrogate model that maps factors to predicted probabilities. Figure~\ref{fig:sample} shows and example of the local linear approximation from an IMDB review classification. %

In high-stakes applications, such as medical diagnosis, it may be risky to rely on a single, noisy prediction from an language model. LAMP addresses this issue by extracting a surrogate model that captures the language model's behavior in a local neighborhood. This allows practitioners to inspect, validate, and potentially adjust model behavior, rather than blindly trusting the language model output. %

Our empirical analyses span tasks including sentiment analysis, controversial-prompt detection, and harmful response detection. Across these tasks, we find that language models produce outputs which lie on a decision surface that we can approximate locally, and we provide a method to establish the radius of approximation. Furthermore, surrogate model coefficients are not in complete alignment with human judgment and expert assessment, reinforcing the practical relevance of LAMP as a tool for expert validation of language model outputs.

\section{RELATED WORKS}

\paragraph{Stochasticity, stability, and robustness of language model outputs}

language models are known to exhibit substantial variability in output, even when queried with the identical inputs and decoding settings. Recent work by \citet{atil25} has quantified the non-deterministic and non-Gaussian behaviors in output generation. Complementary approaches such as Distribution-Based Perturbation Analysis (DBPA) \citet{rauba24a} treat these fluctuations within a frequentist hypothesis test setup. \citet{chen2023, tanneru2024} both use a perturbation-based technique to gain confidence scores for black-box language model responses. This body of work motivates the use of statistical and perturbation-based frameworks for LAMP, which directly utilizes this perspective by estimating local linear decision surfaces via controlled perturbations in explanation space.

\paragraph{Interpretability methods}
Interpretability of black-box models has been a key challenge and an active field of research. LIME \citep{lime} is a popular method that estimates local feature importance via perturbation and  fitting a sparse linear model. On the other hand, SHAP \citep{lundberg17} explains individual predictions by attributing each feature's contribution based on Shapley values from game theory. Aside from perturbation-based methods, there are approaches that aim to be explainable. From Shapely values and surrogate models, partial dependence plots can be constructed to assess linearity \citep{pdp}. Additionally, ANCHORS \citep{ribeiro18} explains individual predictions by identifying high-precision if-then rules that guarantee consistent predictions when certain feature conditions are met. In-Context Explainer (ICE) \citep{kroeger24} directly leverages language model's In-Context Learning (ICL) capabilities to explain the predictions from other models, using language models as explainers without training or architectural changes. Unlike LAMP, ICE targets external model interpretation through free-form explanations, whereas LAMP places emphasis on structured factor representations, and quantitatively measurable self-consistency under input perturbations. Yet, in parallel with these works, LAMP bridges the gap between the perturb-fit approach and the explanation approaches.

\paragraph{Linearity in language model representations}
While large language models are built upon non-linear architectures with billions parameters, a growing number of research papers have theoretically and empirically uncovered linear behaviors in language model internal representations. \citet{tigges23} finds that sentiment is encoded along a single linear direction in the activation space. \citet{nanda23} shows that transformer models can represent its world model linearly. \citet{jiang24} theoretically and empirically shows that the next token prediction objective in language models promotes a linear representation of concepts through a latent variable model, and \citet{park24} provides causal analysis. These studies suggest that, despite the complex nature of language model architectures, local linear approximations are capable of capturing meaningful insights of their behavior. Unlike prior work on latent states or internal architecture, LAMP focuses on externally observable behavior, providing a lightweight and model-agnostic auditing tool for Language Models.

\section{LOCAL ATTRIBUTION MAPPING PROBE (LAMP)}
\label{sec:lamp-parody}
\begin{figure*}[htp]
    \centering
    \includegraphics[width=0.8\linewidth]{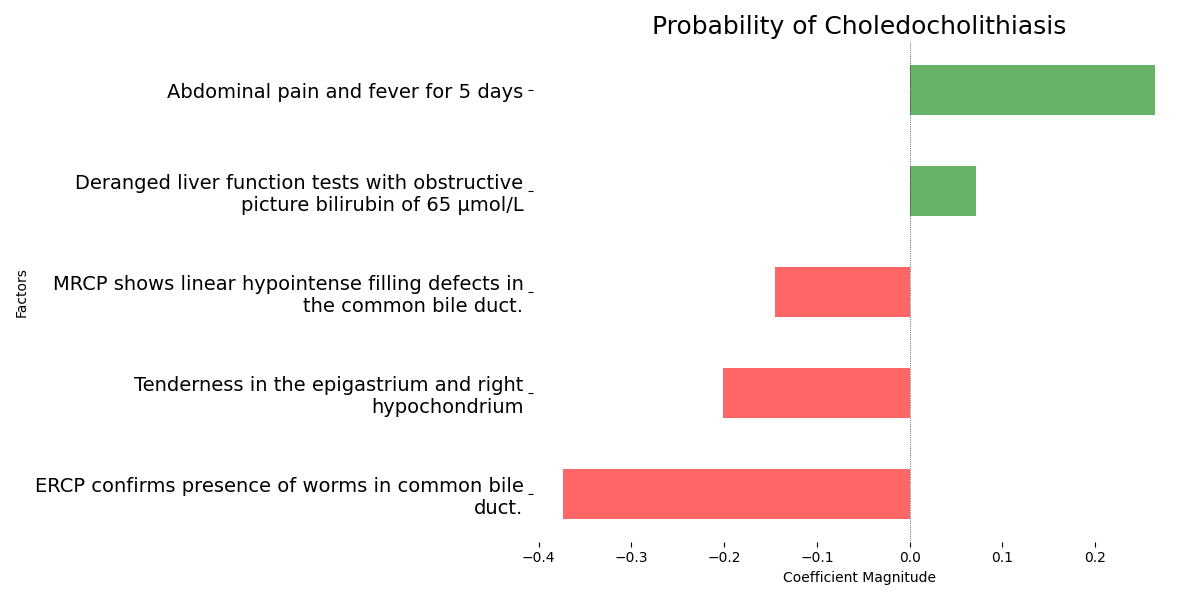}
    \caption{\textbf{LAMP surrogate coefficients visualized for a representative patient case.} LAMP outputs a diagnosis along with a rationale. The surrogate model provides a local approximation of the decision surface, with coefficient magnitudes indicating the direction and relative influence of each factor on the language model's reported probability of diagnosis.}
    \label{fig:coef-example}
\end{figure*}

Imagine a prospective home-buyer who is looking at a house and is deciding on whether to place an offer on it. We may ask the home-buyer to list her considerations such as purchase price, size of backyard, or school district quality. This home buyer will assign different importance weights to each criteria and her agent can ask how likely she will make an offer. 

Now imagine 50 clones of the same home-buyer who are all exactly identical except that they are instantiated with a slight difference in preferences. Perhaps one clone cares slightly less about commute time, and another may care more about the purchase price. All of these clones are presented with the same house and asked to tell how likely they are to make an offer on the house. Their likelihood of buying plausibly would change as preferences shift.

Collecting the clones' 50 decisions and their preferences provide us with a data set $\{\mathbf{w}_i, p_i\}_{i=1}^{50}$, where $\mathbf{w}_i$ forms the importance weights that clone $i$ places on her preferences, and $p_i$ the reported probability of placing an offer that she reports. Because the clones' weights are only slightly perturbed from the original home-buyer, the data set provides signal on the local structure of the home-buyer's decision surface on whether to place an offer on the house. We approximate the decision surface locally with a linear model:
\[
p = \beta^T \mathbf{w} + \varepsilon
\]
such that the coefficients $\beta^T$ tell how strongly and in which direction, each criteria can influence the home-buyer's decision, at least for preferences near or at the buyer's original preference point. The linear model does not estimate the true cognitive processes of the buyer, but rather represents the first order approximation to the home-buyer's decision surface. That is, the linear model reveals which reported ``dials'' the model is responsive to under perturbation, but should not be used to explain the underlying mechanisms that change the buyer's likelihood.

LAMP applies the same logic to large language models. When a model is asked to classify a piece of text, it is also asked to provide the prediction probabilities and importance weights to the list of explanatory factors, the analog to a house's square footage or commute time. LAMP applies perturbations to the importance weights and re-queries the model for probabilities several times, an analog to the 50 identical clones. The weights and probabilities are collected, and a surrogate linear model is fit on the data to tell a story about which dials to turn and by how much, such that we can nudge the model's self-reported labels and probabilities.

\section{METHODOLOGY}
LAMP is a method for probing how a language model's self-explanations relate to its output decisions. We start by outlining how the model is prompted, then how we perturb its self-reported weights to build a local surrogate model, and explain how we determine a suitable scale for the perturbations based on local curvature.

\subsection{Extracting model explanations}
\label{subsec:method-extract}

We begin with a classification input $x$, which language model $\mathcal{M}$ must assign to one of $C$ classes. As part of its response, $\mathcal{M}$ is prompted to provide both the classification probability $\mathbb{P}(x = c \in C)$ and the explanation/reasoning for the classification. The explanation contains its rationales, or factors $\mathbf{f} = f_1,...,f_n$, as well as importance weights $\mathbf{w}_0 = w_1,...,w_n$, denoting what it believes to be the most important factors in determining the classification.

We treat the classification probability as a sample from an unknown function $\Phi(\mathbf{w})+\epsilon$, where $\epsilon$ is a random noise and $\Phi(\cdot)$ is the decision surface mapping factor weights to output probability. This set-up helps us approximate $\Phi(\cdot)$ around the neighborhood of the model's rationale. 

\subsection{Probing the decision surface}
\label{subsec:method-lamp}

To explore how the model's predictions change in response to its self-reported factors, we perturb the weight vector $\mathbf{w}_0$ by adding a stochastic jitter $\delta$. For each perturbed weight vector $\mathbf{w}_0+\delta$, we prompt $\mathcal{M}$ again with the same factors but updated weights, and record the new predicted probability $\mathbb{P}(x = c\in C)$. This probability can be seen as a sample coming from $\Phi(\mathbf{w}_0 + \delta)$ with noise. Repeating this perturbation $m$ times gives us a set of perturbed weights and corresponding probabilities, i.e. a design matrix of rationale weights, $X \in \mathbb{R}^{m\times n}$, and a vector of probabilities, $y \in \mathbb{R}^{m}$. This data provides information about the decision surface, $\Phi$, around the neighborhood of $\mathbf{w}_0$. In partiuclar, to approximate the local geometry of $\Phi$, we fit a linear surrogate model of the form:
\begin{equation}
    \hat y = X\hat\beta,
\end{equation}
where $\hat\beta\in\mathbb{R}^n$ captures the local linear relationship between the factor weights and predicted probability. This surrogate serves as a faithful local proxy for the model's behavior in the neighborhood of $\mathbf{w}_0$. Algorithm \ref{alg:lamp} details the workflow of LAMP.

\RestyleAlgo{ruled}
\begin{algorithm}[htp]
\caption{\textbf{LAMP}: Local Attribution Mapping Probe}
\label{alg:lamp}
\DontPrintSemicolon
\KwIn{Instance $x$ (text), black‑box language model $\mathcal{M}$, 
number of perturbations $m$, 
perturbation scale $\sigma$, 
ridge parameter $\lambda$}
\KwOut{Set of surrogate coefficients $\{\boldsymbol\beta^{(r)}\}_{r=1}^{R}$}

\vspace{2pt}
  {\bf Step 1: self explanation}\;
  \quad Query $\mathcal{M}$ with prompt $\mathsf{Explain}(x)$\;
  \quad Receive probability prediction $y^{(0)}$ and factor weight list 
        $\{(f_i,w_i)\}_{i=1}^{d}$\;
  \quad Store weight vector $\mathbf{w} \leftarrow (w_1,\dots,w_d)$\;
  \quad Store factors $\mathbf{f} \leftarrow (f_1,\dots,f_d)$\;

  \vspace{2pt}
  {\bf Step 2: perturb weight space}\;
  \quad \For{$j \leftarrow 1$ \KwTo $m$}{
    Draw noise $\boldsymbol\epsilon^{(j)} \sim \mathcal{U}(-\delta, \delta)$\;
    $\tilde{\mathbf{w}}^{(j)} \leftarrow \mathbf{w} \odot (1+\boldsymbol\epsilon^{(j)})$\;
    Send $\tilde{\mathbf{w}}^{(j)}$ back to $\mathcal{M}$ with prompt $\mathsf{Relabel}(x,\mathbf{f},\tilde{\mathbf{w}}^{(j)})$\;
    Receive updated probability prediction $y^{(j)}$\;
  }

  \vspace{2pt}
  {\bf Step 3: fit local surrogate}\;
  \quad Let $X \! \in\! \mathbb{R}^{m\times d}$ have rows $\tilde{\mathbf{w}}^{(j)}$; 
        $\mathbf{y} \!\leftarrow\! (y^{(0)},\dots,y^{(m)})$\;
  \quad $\boldsymbol\beta^{(r)} \leftarrow 
         \arg\min_{\boldsymbol\beta} 
         \lVert X\boldsymbol\beta - \mathbf{y}\rVert_2^{2}$\;

\end{algorithm}

\subsection{Determining numbers and dimensionality reduction}
\label{subsec:meta-agg}

We query $\mathcal{M}$ to provide its factors for classification of text $x$, which often produces long lists of explanatory factors. Each factor introduces a new dimension in the explanation space, and this dimensionality leads to challenges with stability and interpretability of the local surrogate models. An analog to the human example in Section \ref{sec:lamp-parody} is typically when a person weighs factors to make a decision, it is hard for her to juggle several factors at once. Instead, she may rely on a smaller number of factors to help make her decision \citep{gigerenzer2011heuristic}. Statistically speaking, fitting a stable linear approximation over many dimensions requires more perturbation data, and many of the long-tail factors that contribute little to the model's actual decision surface, but may correlate highly with other factors, inflate variances. 

To address this, we introduce a meta-aggregation step designed to reduce dimensionality while preserving usefulness as a factor. We prompt several rationales from $\mathcal{M}$ by repeatedly requesting self-explanations. With the aggregated explanation list, we independently prompt $\mathcal{M}$ again to summarize and consolidate the explanations into a smaller set of $n$  factors, each representing an aggregate theme from the original list. This procedure can be viewed as a method of unsupervised feature selection \citep{jeong2024llmselect, li2025exploring}. However, to ensure that this model selection produces effective predictors for the surrogate model, we compare two versions of LAMP: one with the full set of rationales, and another with the aggregated set. We fit surrogates with and without this aggregation step in Appendix~\ref{app:meta-agg} and show that the aggregation generates the preferred set of factors. From Appendix~\ref{app:meta-agg}, we opt to use $n=5$ aggregated rationales to reduce model complexity and promote interpretability.

\subsection{Choosing the perturbation scale}
\label{app:method-scale}

A central challenge in constructing local linear approximations is determining an appropriate perturbation parameter $\delta$. The goal is to identify a neighborhood around a reference point $\mathbf{w}_0$ in which the decision surface of $\Phi(\mathbf{w}_0 + \delta)$ (maximally perturbed) remains sufficiently linear to justify a first-order linear approximation. Since $\Phi$ is unknown, we adopt a second-order Taylor expansion approximation:
\begin{equation}
\label{eqn:taylor}
    \Phi(\textbf{w}_0 + \boldsymbol{\delta}) \approx \Phi(\textbf{w}_0) + \nabla\Phi(\mathbf{w}_0) \boldsymbol{\delta} + \frac{1}{2}\boldsymbol{\delta}^T H\boldsymbol{\delta}.
\end{equation}
The linear surrogate corresponds to the first-order term, while the second-order term introduces curvature governed by the local Hessian $H$. 

In the absence of a parametric form for $\Phi$, we estimate local curvature empirically by regressing model outputs on perturbed inputs using a second-order model:
\begin{equation}
    y_i = \beta^T \delta_i + \delta_i^T H \delta_i + \varepsilon_i.
\end{equation}
This formulation captures both linear and quadratic effects, yielding estimates $\hat{\beta}$ and $\hat{H}$. From the Taylor approximation, we derive the mean squared error (MSE) of the surrogate model as a function of $\delta$, the parameter of the uniform kernel of $\mathcal{U}(-\delta, \delta)$. We arrive at the MSE:
\begin{equation}
\label{eqn:mse}
MSE(\delta) = \frac{1}{36}\|H\|_F^2\delta^4 + \frac{\sigma^2}{n\delta^d}.
\end{equation}
The derivation for \ref{eqn:mse} is deferred to \ref{app:perturb-ablation}. Here, $d$ is the input dimensionality, $\sigma^2$ is the variance of the residuals. The optimal perturbation radius minimizing this MSE is:
\begin{equation}
\label{eqn:optimal-perturb}
    \delta^* = \left(\frac{9d\hat{\sigma}^2}{n \|H\|_F^2}\right)^{\frac{1}{4+d}}.
\end{equation}

The expression in Equation \ref{eqn:optimal-perturb} indicates that the optimal perturbation radius decreases with increasing curvature and sample size. Since $\delta^*$ must be estimated from sampled data, we adopt a post hoc procedure: if the perturbation radius used exceeds $\delta^*$, we discard any perturbations with $\delta > \delta^*$, thereby ensuring that the surrogate remains within a regime where the linear approximation is valid.

\section{EXPERIMENTS}

In this section, we systematically evaluate LAMP's capability to approximate the decision surfaces of large language models through locally linear surrogate models derived from the language models' own explanations. Our experiments are structured around three core objectives: assessing the quality of the linear fit (Section \ref{sec:r2}), assessing the consistency of surrogate model predictions with actual language model outputs (Section \ref{sec:prediction}), and evaluating whether LAMP can produce meaningful cues that can inform human decisions (Section \ref{sec:human}). Together, these experiments validate the hypothesis that language model decision surfaces exhibit local linear structure that can be effectively be a useful tool to characterize and explain a model's decision landscape.

We conduct experiments on three publicly available, binary-label corpora: IMDB reviews \citep{imdb-ds}, Pseudo-Harmful (PH) \citep{phtest}, and HateBenchSet (HateBS) \citep{hatebenchset}. We include a brief description of the data sets and their zero-shot classification performance in Appendix~\ref{app:data-sets}. All three data sets are binary classification problems. In addition, we evaluate a multiclass classification dataset consisting of clinical case studies with different diagnoses \citet{ko2025automated} in the domain of gastroenterology. This dataset was annotated by three gastroenterologists and is included to demonstrate the utility of LAMP in supporting expert human evaluation, particularly in high-stakes and domain-specific settings (Appendix~\ref{app:data-sets}).

\subsection{Explained proportion of variance}
\label{sec:r2}
We begin by examining how well the linear surrogates capture the behavior of language models in their explanation space. The coefficient of determination, $R^2$, provides a standard way of evaluating this, measuring how much of the variance in predicted probabilities is explained by the linear surrogate. 

Table~\ref{tab:r2-main} presents the average $R^2$ values for surrogate models fitted by LAMP. LAMP surrogate $R^2$ values are comparable to LIME surrogate $R^2$ values presented in Appendix~\ref{app:lime-eg}. However, LAMP features are more interpretable as exemplified in Figure~\ref{fig:coef-example}, because LAMP forms features in the explanation space whereas LIME is most typically applied in the token space, which has limited utility as demonstrated in Appendix~\ref{app:lime-eg}. The ability to generate natural language explanation from LAMP features prove to be a useful tool in downstream tasks such as in expert medical diagnosis, discussed in Section \ref{sec:human}.

By construction, $R^2$ is weakly monotone increasing in the number of predictors: removing predictors can only weakly decrease $R^2$. The $R^2$ values in Table~\ref{tab:r2-main} are also comparable to those obtained by best-subset selection from the full pool of factors, which we present in Appendix~\ref{app:meta-agg}. Moreover, we note an interesting observation, that regions where the model is less certain (mid-range probabilities), surrogates achieve higher $R^2$ and exhibit larger coefficient norms. We further explore this phenomenon in Appendix~\ref{app:tail}. Consequently, the modest $R^2$ values in Table~\ref{tab:r2-main} are best interpreted not as evidence against model misspecification, but as a reflection of the decision surface's geometry, that is steep in the middle and flat in the tails.

\begin{table}[ht]
    \centering
    \caption{\textbf{Coefficient of determination ($R^2$) values for local linear surrogate models.} Higher $R^2$ indicates more variability explainable by the surrogate model. Lower values on HateBenchSet and tail regions reflect the reduced variance explained discussed in Appendix~\ref{app:tail}. For reference, the best subset of predictors produce $R^2$ values presented in Appendix~\ref{app:meta-agg}.}
    \resizebox{\linewidth}{!}{%
    \begin{tabular}{lccc}
        \toprule
        \textbf{Model} & IMDB & PH & HateBS \\
        \midrule
        GPT-4.1-mini     & $0.42 \pm 0.03$ & $0.38 \pm 0.03$ & $0.17 \pm 0.02$ \\
        Gemini 2.5 Flash & $0.26 \pm 0.03$ & $0.25 \pm 0.03$ & $0.23 \pm 0.03$ \\
        Claude 3.5 Haiku & $0.26 \pm 0.03$ & $0.21 \pm 0.02$ & $0.16 \pm 0.01$ \\
        Mistral Large    & $0.30 \pm 0.03$ & $0.25 \pm 0.03$ & $0.20 \pm 0.02$ \\
        \bottomrule
    \end{tabular}%
    }
    \label{tab:r2-main}
\end{table}

\subsection{Consistency of surrogate models predictions}
\label{sec:prediction}

In Algorithm \ref{alg:lamp}, we evaluated the language model's self-generated counterfactual hypotheses by instructing it to perturb the importance weights of specific factors within a local region $\delta$ and then asking the language model to produce a new classification and probability. While these perturbations allow us to probe the local decision surface, it is not clear that the perturbation-assigned factor weights translate into corresponding changes in the language model's predictive behavior when using the same perturbations that are instantiated in natural language.

\begin{figure*}[htp!]
    \centering
    \includegraphics[width=0.9\linewidth]{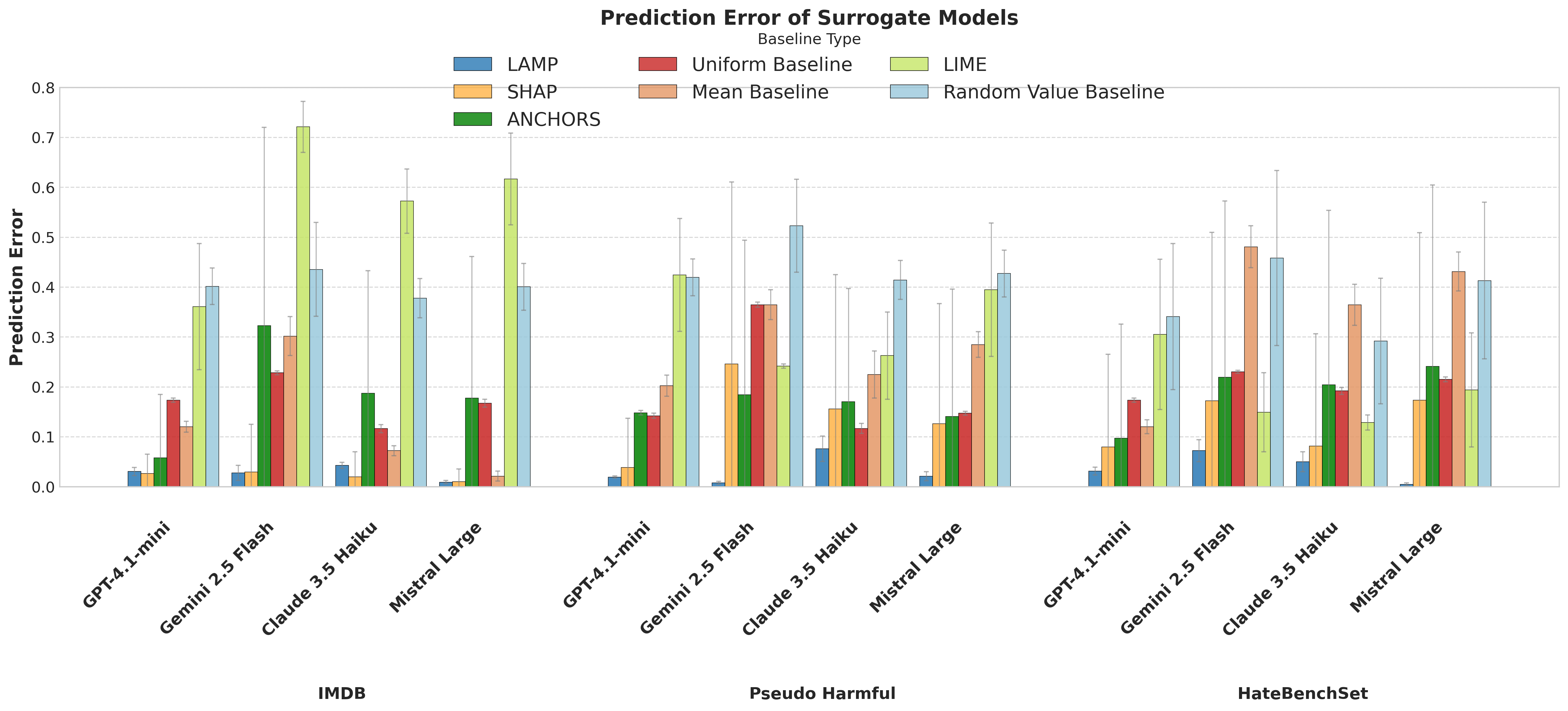}
    \caption{\textbf{Across the board, LAMP surrogate linear models are able to predict out-of-sample language model output to a small margin of error.} GPT, Gemini, and Mistral are able to predict the language model given probability just by using its locally linear surrogate (lower is better). This is better than using an intercept only surrogate (mean model), a naive baseline (always predicting 0.5), and a random selection baseline (predicting 0 or 1 at random). In some settings, SHAP surrogate models are competitive with LAMP surrogate models. A table to show the numbers clearly is available in Appendix~\ref{app:all-errors}.}
    \label{fig:all-errors}
\end{figure*}

Thus to assess the validity of this counterfactual analysis, we will take the original text $x$ and ask an language model to rewrite it into $x_{\text{modified}}$, creating a modified version of the document that encodes different factor weights. Figure~\ref{fig:rewrite-side} provides a brief example of such rewritten text. In Appendix~\ref{app:rewritten}, we show that the perturbations and rewritings preserve both the semantic similarity between $x_{\text{modified}}$ and $x$, and that the factor scores extracted from the rewritten inputs remain close to those of the original.

As the $x_{\text{modified}}$ can be treated as locally perturbed in the factor space, we can then use $x_{\text{modified}}$ to see if the LAMP surrogate linear model can predict the language model given probability. For each $x_{\text{modified}}$, we prompt the language model $\mathcal{M}$ to return a class probability $p_h$ as well as factor weights $w_h$. The factor weights $w_h$ are then passed through the surrogate to get an estimated $\hat{p}_s$. We then measure the forecast error with the Brier score $s(p_h,\hat{p}_s) = (p_h - \hat{p}_s)^2$ \citep{brier} to assess the quality of prediction. Within our classification suite, a perfect prediction corresponds to a Brier score of 0, while the worst prediction yields a Brier score of 1. We evaluate the predictive ability with two baselines: a uniform baseline, where a naive surrogate model predicts 0.5 all the time, and an intercept only baseline, where we take the mean LAMP surrogate prediction, that is, the intercept only model.

\begin{figure}[htp]
\begin{subfigure}[t]{0.45\textwidth}
\begin{tcolorbox}[title=Original $x$, colback=gray!5, colframe=gray!50!black, fonttitle=\bfseries]
Dumb is as dumb does, in this \remove{thoroughly uninteresting}, supposed black comedy [\dots] of ""The Three Amigos"", \remove{only without any laughs}. [\dots] for black comedy to work, it cannot be \remove{mean spirited}, which ""Play Dead"" \remove{is}. What ""Play Dead"" really is, is a town \remove{full of nut jobs}. Fred Dunst does however do a [\dots].
\end{tcolorbox}
\end{subfigure}
\hfill
\begin{subfigure}[t]{0.45\textwidth}
\begin{tcolorbox}[title=Modified $x_{\text{modified}}$, colback=gray!5, colframe=gray!50!black, fonttitle=\bfseries]
Dumb is as dumb does, in this \possent{somewhat engaging}, supposed black comedy [\dots] of ""The Three Amigos"", \negsent{without any substantive comedic relief whatsoever}. [\dots] for black comedy to work, it cannot be \negsent{excessively critical}, which ""Play Dead"" \possent{might be}. What ""Play Dead"" really is, is a town \possent{brimming with eccentric and unhinged characters}. Fred Dunst does however do a [\dots].
\end{tcolorbox}
\end{subfigure}
\caption{\textbf{Example of rewriting a LAMP factors-based perturbation}. In the original text, LAMP identified the blue highlighted text as influential to its classification decision. We perturb their importance scores and rewrite $x$ to $x_{\text{modified}}$. Green highlights are rewritings with a more positive perturbation, while red rewritings suggest a more negative perturbation.}
\label{fig:rewrite-side}
\end{figure}

As seen in Figure~\ref{fig:all-errors}, across the board, the LAMP surrogate model has a relatively low prediction error; further discussion of baseline comparisons, along with the exact error values, is provided in Appendix~\ref{app:all-errors}. Also shown in the figure is that LIME surrogate models poorly capture the local decision surface, even relative to their own self-reported factors and predicted probabilities, suggesting that LIME's features fail to approximate the decision boundary as well as LAMP. This is consistent with prior evidence that LIME's token-masking perturbations are often not truly ``local'' \citep{tan2023glime, zafar2019dlime}, and in some cases LIME performs worse than naive baselines. For SHAP, we observe two IMDB settings (GPT-4.1-mini and Claude~3.5~Haiku) where SHAP attains a marginally lower prediction error than LAMP. However, SHAP's prediction errors have substantially higher variance, and across the remaining experiments in Figure~\ref{fig:all-errors}, LAMP achieves both lower error and markedly lower variance.
In addition to black-box perturbation methods, we also compare LAMP to Integrated Gradients, a white-box approach requiring access to gradients and token-level logits (Appendix~\ref{app:white-box})

With win-rate/average-rank summaries and confidence intervals taken together, these results indicate that LAMP produces more reliable surrogates overall, while the isolated SHAP wins are not statistically significant given their variability.
This suggests that the LAMP surrogate is valid via construction by counterfactual generations, and the LAMP surrogate models are a good local approximation to the language model decision surface. Furthermore, this suggests that language models exhibit a degree of self-consistency. Their reported probabilities are correlated with the directions indicated by their own reported factors and importance weights. In addition to the prediction error, we demonstrate the correlation between the predicted probabilities and the language model reported probabilities in Appendix~\ref{app:prediction-corr}.

\subsection{Evaluating the interpretive validity and usefulness of LAMP}
\label{sec:human}

\begin{table*}[h!]
    \centering
    \label{tab:human-agree}
    \caption{\textbf{Evaluator agreement with LAMP factor attributions.} Agreement between surrogate model factor weights and expected feature directions across datasets. Higher values indicate stronger correspondence between surrogate attributions and anticipated input-output relationships. GPT-4.1-mini consistently exhibits patterns that are more aligned with these reference expectations. There were 3 gastroenterology experts who rated the factors, and their interagreement score is 0.635.}
    \resizebox{\linewidth}{!}{
    \begin{tabular}{lcccc}
        \toprule
        \textbf{Dataset} & \textbf{GPT-4.1-mini} & \textbf{Gemini 2.5 Flash} & \textbf{Claude 3.5 Haiku} & \textbf{Mistral Large} \\
        \midrule
        IMDB              & 0.840 & 0.700 & 0.620 & 0.680    \\
        Pseudo Harmful    & 0.780 & 0.729 & 0.600 & 0.660 \\
        HateBenchSet      & 0.600 & 0.633 & 0.680 & 0.640 \\
        Gastroenterology (0.635)  & 0.697 & 0.624 & 0.645 & 0.461 \\
        \bottomrule
    \end{tabular}}
\end{table*}

LAMP offers a practical tool for users to investigate and characterize the decision behavior of large language models. LAMP allows users to pose the question ``Why should I trust you?'' to a language model \citep{lime}, and obtain interpretable rationales along with a local approximation of the model's decision surface.

We conducted a qualitative analysis of LAMP surrogate model coefficients to assess whether these local linear approximations capture interpretable patterns in the model's decision surface. For each input text, we examined the directionality of the surrogate features and compared them to expectations about how such features might influence classification outcomes. In the gastroenterology dataset, which involves specialized domain knowledge, directional assessments were evaluated by 3 physicians with specialty training in gastroenterology, averaging their agreement scores.

By comparing these directional expectations with the surrogate coefficients, we assess whether the resulting linear trends reflect patterns that could plausibly be used to inspect or reason about model behavior. Strong correspondence suggests that the surrogate surfaces may encode structure consistent with domain-relevant signals, potentially aiding in interpretability. Weaker correspondence, conversely, may indicate that the model's reported decision factors lack coherence with known or expected input-output relationships.

Among the models examined, Claude shows the strongest correspondence with expectations on the HateBenchSet dataset, which aligns with its popularly known tendency toward safety-aligned outputs. Mistral, by contrast, shows relatively weaker correspondence, while GPT models tend to exhibit more consistent patterns across three of the data sets. These observations point to differences in how clearly each model's reported decision factors align with patterns of behavior, potentially informing when and how their outputs might be considered more interpretable or reliable.

We particularly note that agreement with physicians is relatively low across the board. However, the inter-physician agreement is 0.635. Since both GPT agreement and Claude agreement both score higher than the inter-physician agreement, this suggests two things. First, the model's factors and coefficients align more closely with each physician individually than the physicians align with each other. Additionally, the model's outputs are not random, rather, they capture a signal that is at least as factually valid as the physicians' own judgments. These results highlight LAMP's potential as a decision support tool by providing structured explanations that remain accessible and reviewable, even when human judgment may vary.

\section{DISCUSSION AND CONCLUSION}

In this work, we test whether a language model's self-declared factors form a behaviorally coherent decision surface and can define a local approximation of the decision surface. We also evaluate LAMP as a tool that can help users assess the validity of a language model's decisions. Section \ref{sec:r2} show that LAMP surrogate models produce approximations to the decision surface, and Section \ref{sec:prediction} show that LAMP surrogate models built from stochastic weight jitters anticipate the probabilities produced by natural-language counterfactuals, confirming that the self-reported factors can act as bona-fide control dials.

LAMP is a useful tool to audit language model decisions to promote interpretability, which can be helpful to promote trust in scenarios where human-algorithmic interactions are important for making complex decisions. This is particularly true in clinical applications such as diagnostic reasoning, where language models show promise (\cite{liu2025generalist}, \cite{tu2025conversational}). LAMP could be deployed in situations where language models are used to assist physicians with diagnostic reasoning.

Looking forward, we plan to lift the linearity assumption by seeding higher-order surrogates with the same curvature estimates that govern the LAMP radius, to generalize the probe beyond classification to generative and sequential settings, and to embed it in interactive audit tools that let practitioners steer model outputs in real time. These extensions aim to turn the behavioral regularities revealed here into actionable transparency for increasingly capable, increasingly opaque language models.

\newpage
\bibliography{main}

@inproceedings{faith-hard, address={Online}, title={Towards Faithfully Interpretable NLP Systems: How Should We Define and Evaluate Faithfulness?}, url={https://www.aclweb.org/anthology/2020.acl-main.386}, DOI={10.18653/v1/2020.acl-main.386}, booktitle={Proceedings of the 58th Annual Meeting of the Association for Computational Linguistics}, publisher={Association for Computational Linguistics}, author={Jacovi, Alon and Goldberg, Yoav}, year={2020}, pages={4198–4205}, language={en} }

@inproceedings{faith-or-sc, address={Bangkok, Thailand}, title={On Measuring Faithfulness or Self-consistency of Natural Language Explanations}, url={https://aclanthology.org/2024.acl-long.329}, DOI={10.18653/v1/2024.acl-long.329}, booktitle={Proceedings of the 62nd Annual Meeting of the Association for Computational Linguistics (Volume 1: Long Papers)}, publisher={Association for Computational Linguistics}, author={Parcalabescu, Letitia and Frank, Anette}, year={2024}, pages={6048–6089}, language={en} }

@inproceedings{lime, address={San Francisco California USA}, title={“Why Should I Trust You?”: Explaining the Predictions of Any Classifier}, ISBN={9781450342322}, url={https://dl.acm.org/doi/10.1145/2939672.2939778}, DOI={10.1145/2939672.2939778}, booktitle={Proceedings of the 22nd ACM SIGKDD International Conference on Knowledge Discovery and Data Mining}, publisher={ACM}, author={Ribeiro, Marco Tulio and Singh, Sameer and Guestrin, Carlos}, year={2016}, month=aug, pages={1135–1144}, language={en} }

@article{sc-cot,
  title={Self-consistency improves chain of thought reasoning in language models},
  author={Wang, Xuezhi and Wei, Jason and Schuurmans, Dale and Le, Quoc and Chi, Ed and Narang, Sharan and Chowdhery, Aakanksha and Zhou, Denny},
  journal={arXiv preprint arXiv:2203.11171},
  year={2022}
}

@InProceedings{imdb-ds,
  author    = {Maas, Andrew L.  and  Daly, Raymond E.  and  Pham, Peter T.  and  Huang, Dan  and  Ng, Andrew Y.  and  Potts, Christopher},
  title     = {Learning Word Vectors for Sentiment Analysis},
  booktitle = {Proceedings of the 49th Annual Meeting of the Association for Computational Linguistics: Human Language Technologies},
  month     = {June},
  year      = {2011},
  address   = {Portland, Oregon, USA},
  publisher = {Association for Computational Linguistics},
  pages     = {142--150},
  url       = {http://www.aclweb.org/anthology/P11-1015}
}

@inproceedings{hatebenchset,
  author = {Xinyue Shen and Yixin Wu and Yiting Qu and Michael Backes and Savvas Zannettou and Yang Zhang},
  title = {{HateBench: Benchmarking Hate Speech Detectors on LLM-Generated Content and Hate Campaigns}},
  booktitle = {{USENIX Security Symposium (USENIX Security)}},
  publisher = {USENIX},
  year = {2025}
}

@inproceedings{
phtest,
title={Automatic Pseudo-Harmful Prompt Generation for Evaluating False Refusals in Large Language Models},
author={Bang An and Sicheng Zhu and Ruiyi Zhang and Michael-Andrei Panaitescu-Liess and Yuancheng Xu and Furong Huang},
booktitle={First Conference on Language Modeling},
year={2024},
url={https://openreview.net/forum?id=ljFgX6A8NL}
}

@inproceedings{
cot,
title={Chain-of-Thought Prompting Elicits Reasoning in Large Language Models},
author={Jason Wei and Xuezhi Wang and Dale Schuurmans and Maarten Bosma and brian ichter and Fei Xia and Ed H. Chi and Quoc V Le and Denny Zhou},
booktitle={Advances in Neural Information Processing Systems},
editor={Alice H. Oh and Alekh Agarwal and Danielle Belgrave and Kyunghyun Cho},
year={2022},
url={https://openreview.net/forum?id=_VjQlMeSB_J}
}

@inproceedings{
world-reps,
title={Do {LLM}s Build World Representations? Probing Through the Lens of State Abstraction},
author={Zichao Li and Yanshuai Cao and Jackie CK Cheung},
booktitle={The Thirty-eighth Annual Conference on Neural Information Processing Systems},
year={2024},
url={https://openreview.net/forum?id=lzfzjYuWgY}
}

@article{guesstimate,
  title={Probing LLM World Models: Enhancing Guesstimation with Wisdom of Crowds Decoding},
  author={Chuang, Yun-Shiuan and Harlalka, Nikunj and Narendran, Sameer and Cheung, Alexander and Gao, Sizhe and Suresh, Siddharth and Hu, Junjie and Rogers, Timothy T},
  journal={arXiv preprint arXiv:2501.17310},
  year={2025}
}

@inproceedings{
turpin2023,
title={Language Models Don't Always Say What They Think: Unfaithful Explanations in Chain-of-Thought Prompting},
author={Miles Turpin and Julian Michael and Ethan Perez and Samuel R. Bowman},
booktitle={Thirty-seventh Conference on Neural Information Processing Systems},
year={2023},
url={https://openreview.net/forum?id=bzs4uPLXvi}
}

@inproceedings{webson2022,
  title={Do prompt-based models really understand the meaning of their prompts?},
  author={Webson, Albert and Pavlick, Ellie},
  booktitle={Proceedings of the 2022 Conference of the North American Chapter of the Association for Computational Linguistics: Human Language Technologies},
  pages={2300--2344},
  year={2022}
}

@article{min2022,
  title={Rethinking the role of demonstrations: What makes in-context learning work?},
  author={Min, Sewon and Lyu, Xinxi and Holtzman, Ari and Artetxe, Mikel and Lewis, Mike and Hajishirzi, Hannaneh and Zettlemoyer, Luke},
  journal={arXiv preprint arXiv:2202.12837},
  year={2022}
}

@misc{atil25,
  title = {Non-{{Determinism}} of "{{Deterministic}}" {{LLM Settings}}},
  author = {Atil, Berk and Aykent, Sarp and Chittams, Alexa and Fu, Lisheng and Passonneau, Rebecca J. and Radcliffe, Evan and Rajagopal, Guru Rajan and Sloan, Adam and Tudrej, Tomasz and Ture, Ferhan and Wu, Zhe and Xu, Lixinyu and Baldwin, Breck},
  year = {2025},
  month = apr,
  number = {arXiv:2408.04667},
  eprint = {2408.04667},
  primaryclass = {cs},
  publisher = {arXiv},
  doi = {10.48550/arXiv.2408.04667},
  urldate = {2025-05-12}
}

@inproceedings{tanneru2024,
  title={Quantifying uncertainty in natural language explanations of large language models},
  author={Tanneru, Sree Harsha and Agarwal, Chirag and Lakkaraju, Himabindu},
  booktitle={International Conference on Artificial Intelligence and Statistics},
  pages={1072--1080},
  year={2024},
  organization={PMLR}
}

@article{chen2023,
  title={Quantifying uncertainty in answers from any language model and enhancing their trustworthiness},
  author={Chen, Jiuhai and Mueller, Jonas},
  journal={arXiv preprint arXiv:2308.16175},
  year={2023}
}

@misc{jiang24,
  title = {On the {{Origins}} of {{Linear Representations}} in {{Large Language Models}}},
  author = {Jiang, Yibo and Rajendran, Goutham and Ravikumar, Pradeep and Aragam, Bryon and Veitch, Victor},
  year = {2024},
  month = mar,
  number = {arXiv:2403.03867},
  eprint = {2403.03867},
  primaryclass = {cs},
  publisher = {arXiv},
  doi = {10.48550/arXiv.2403.03867},
  urldate = {2025-05-12}
}

@misc{kroeger24,
  title = {In-{{Context Explainers}}: {{Harnessing LLMs}} for {{Explaining Black Box Models}}},
  shorttitle = {In-{{Context Explainers}}},
  author = {Kroeger, Nicholas and Ley, Dan and Krishna, Satyapriya and Agarwal, Chirag and Lakkaraju, Himabindu},
  year = {2024},
  month = jul,
  number = {arXiv:2310.05797},
  eprint = {2310.05797},
  primaryclass = {cs},
  publisher = {arXiv},
  doi = {10.48550/arXiv.2310.05797},
  urldate = {2025-05-12}
}

@misc{lundberg17,
  title = {A {{Unified Approach}} to {{Interpreting Model Predictions}}},
  author = {Lundberg, Scott and Lee, Su-In},
  year = {2017},
  month = nov,
  number = {arXiv:1705.07874},
  eprint = {1705.07874},
  primaryclass = {cs},
  publisher = {arXiv},
  doi = {10.48550/arXiv.1705.07874},
  urldate = {2025-05-12}
}

@misc{nanda23,
  title = {Emergent {{Linear Representations}} in {{World Models}} of {{Self-Supervised Sequence Models}}},
  author = {Nanda, Neel and Lee, Andrew and Wattenberg, Martin},
  year = {2023},
  month = sep,
  number = {arXiv:2309.00941},
  eprint = {2309.00941},
  primaryclass = {cs},
  publisher = {arXiv},
  doi = {10.48550/arXiv.2309.00941},
  urldate = {2025-05-12}
}

@misc{park24,
  title = {The {{Linear Representation Hypothesis}} and the {{Geometry}} of {{Large Language Models}}},
  author = {Park, Kiho and Choe, Yo Joong and Veitch, Victor},
  year = {2024},
  month = jul,
  number = {arXiv:2311.03658},
  eprint = {2311.03658},
  primaryclass = {cs},
  publisher = {arXiv},
  doi = {10.48550/arXiv.2311.03658},
  urldate = {2025-05-12},
}

@misc{rauba24a,
  title = {Quantifying Perturbation Impacts for Large Language Models},
  author = {Rauba, Paulius and Wei, Qiyao and van der Schaar, Mihaela},
  year = {2024},
  month = dec,
  number = {arXiv:2412.00868},
  eprint = {2412.00868},
  primaryclass = {cs},
  publisher = {arXiv},
  doi = {10.48550/arXiv.2412.00868},
  urldate = {2025-05-12},
}

@article{ribeiro18,
  title = {Anchors: {{High-Precision Model-Agnostic Explanations}}},
  shorttitle = {Anchors},
  author = {Ribeiro, Marco Tulio and Singh, Sameer and Guestrin, Carlos},
  year = {2018},
  month = apr,
  journal = {Proceedings of the AAAI Conference on Artificial Intelligence},
  volume = {32},
  number = {1},
  issn = {2374-3468, 2159-5399},
  doi = {10.1609/aaai.v32i1.11491},
  urldate = {2025-05-12},
}

@misc{tigges23,
  title = {Linear {{Representations}} of {{Sentiment}} in {{Large Language Models}}},
  author = {Tigges, Curt and Hollinsworth, Oskar John and Geiger, Atticus and Nanda, Neel},
  year = {2023},
  month = oct,
  number = {arXiv:2310.15154},
  eprint = {2310.15154},
  primaryclass = {cs},
  publisher = {arXiv},
  doi = {10.48550/arXiv.2310.15154},
  urldate = {2025-05-12},
}

@article{brier,
  title={Verification of forecasts expressed in terms of probability},
  author={Brier, Glenn W},
  journal={Monthly weather review},
  volume={78},
  number={1},
  pages={1--3},
  year={1950}
}

@article{pdp,
  title={Greedy function approximation: a gradient boosting machine},
  author={Friedman, Jerome H},
  journal={Annals of statistics},
  pages={1189--1232},
  year={2001},
  publisher={JSTOR}
}

@article{harvey1977testing,
  title={Testing for functional misspecification in regression analysis},
  author={Harvey, Andrew C and Collier, Patrick},
  journal={Journal of Econometrics},
  volume={6},
  number={1},
  pages={103--119},
  year={1977},
  publisher={Elsevier}
}

@article{Zimmer2019,
  author    = {Zimmer, V.},
  title     = {Naked Fat Sign Is a Characteristic of Colonic Lipoma},
  journal   = {Clinical Gastroenterology and Hepatology},
  year      = {2019},
  month     = feb,
  volume    = {17},
  number    = {3},
  pages     = {A29},
  doi       = {10.1016/j.cgh.2018.02.046},
  pmid      = {29510213}
}

@article{Chatila2019,
  author    = {Chatila, A. T. and Bilal, M. and Merwat, S.},
  title     = {Kayexalate-Induced Colonic Pseudotumor},
  journal   = {Clinical Gastroenterology and Hepatology},
  year      = {2019},
  month     = jun,
  volume    = {17},
  number    = {7},
  pages     = {e73},
  doi       = {10.1016/j.cgh.2018.03.032},
  pmid      = {29609060}
}

@article{Chudy2019,
  author    = {Chudy-Onwugaje, K. and Vandermeer, F. and Quezada, S.},
  title     = {Mimicking Abdominal Tuberculosis: Abdominal Abscess Caused by Lawsonella clevelandensis in Inflammatory Bowel Disease},
  journal   = {Clinical Gastroenterology and Hepatology},
  year      = {2019},
  month     = jul,
  volume    = {17},
  number    = {8},
  pages     = {e92},
  doi       = {10.1016/j.cgh.2018.06.017},
  pmid      = {29913279},
  pmcid     = {PMC6298851}
}

@article{Noor2020,
  author    = {Mohd Noor, N. A. and Goh, S. N. and Tan, C. H.},
  title     = {Biliary Ascariasis: An Unusual Case of Obstructive Jaundice},
  journal   = {Clinical Gastroenterology and Hepatology},
  year      = {2020},
  month     = jun,
  volume    = {18},
  number    = {7},
  pages     = {A16},
  doi       = {10.1016/j.cgh.2019.01.032},
  pmid      = {30710698}
}

@article{Liang2011,
  author    = {Liang, H. H. and Wei, P. L. and Huang, M. T.},
  title     = {"Rolling ball" in the abdomen. Mesenteric cyst in mesenterium of proximal jejunum},
  journal   = {Gastroenterology},
  year      = {2011},
  month     = mar,
  volume    = {140},
  number    = {3},
  pages     = {e9--e10},
  doi       = {10.1053/j.gastro.2010.03.060},
  pmid      = {21288798}
}

@article{Sweetser2013,
  author    = {Sweetser, S. and Chandan, V. S. and Baron, T. H.},
  title     = {Dysphagia in Lynch syndrome},
  journal   = {Gastroenterology},
  year      = {2013},
  month     = nov,
  volume    = {145},
  number    = {5},
  pages     = {945, 1167--8},
  doi       = {10.1053/j.gastro.2013.08.006},
  pmid      = {24075947}
}

@article{Nunes2014,
  author    = {Nunes, T. and Chagas, Mde S. and Biccas, B.},
  title     = {A 70-year-old woman with dysphagia beginning 6 decades after caustic ingestion},
  journal   = {Gastroenterology},
  year      = {2014},
  month     = may,
  volume    = {146},
  number    = {5},
  pages     = {1174, 1430--1},
  doi       = {10.1053/j.gastro.2014.01.002},
  pmid      = {24680964}
}

@article{Wu2018,
  author    = {Wu, J. and Wang, Y. and Wang, C.},
  title     = {Amyloidosis: An Unusual Cause of Intestinal Pseudo-Obstruction},
  journal   = {Clinical Gastroenterology and Hepatology},
  year      = {2018},
  month     = may,
  volume    = {16},
  number    = {5},
  pages     = {e53--e54},
  doi       = {10.1016/j.cgh.2017.07.002},
  pmid      = {28711693}
}

@article{Iida2018,
  author    = {Iida, T. and Yamashita, K. and Nakase, H.},
  title     = {Localized Gastrointestinal AL$\lambda$ Amyloidosis},
  journal   = {Clinical Gastroenterology and Hepatology},
  year      = {2018},
  month     = sep,
  volume    = {16},
  number    = {9},
  pages     = {e93},
  doi       = {10.1016/j.cgh.2017.09.026},
  pmid      = {29627428}
}

@article{Kothadia2018,
  author    = {Kothadia, J. P. and Kone, V. and Raza, A.},
  title     = {Double Major Duodenal Papillae: A Rare Congenital Anomaly of Hepatic and Pancreatic Drainage System},
  journal   = {Clinical Gastroenterology and Hepatology},
  year      = {2018},
  month     = jul,
  volume    = {16},
  number    = {7},
  pages     = {A39--A40},
  doi       = {10.1016/j.cgh.2017.09.053},
  pmid      = {29421157}
}

@article{Nagashima2018,
  author    = {Nagashima, K. and Katsurada, T. and Sakamoto, N.},
  title     = {A Case of Olmesartan-associated Sprue-like Enteropathy},
  journal   = {Clinical Gastroenterology and Hepatology},
  year      = {2018},
  month     = oct,
  volume    = {16},
  number    = {10},
  pages     = {A45--A46},
  doi       = {10.1016/j.cgh.2017.12.015},
  pmid      = {29247793}
}

@article{Wee2013,
  author    = {Wee, Eric and Buenaseda, Ma Clarissa},
  title     = {Dysphagia Due to a "Freak of Nature"},
  journal   = {Gastroenterology},
  year      = {2013},
  volume    = {144},
  number    = {2},
  pages     = {273},
}

@inproceedings{ko2025automated,
  author    = {Ko, Y. and Yun, N. and Stadie, B. and Shung, D.},
  title     = {Automated Prompt Optimization Strategy Improves Large Language Model Diagnostic Accuracy for Complex Clinical Cases in Gastroenterology and Hepatology},
  booktitle = {Digestive Diseases Week 2025},
  year      = {2025},
  address   = {San Diego, CA},
  note      = {Conference abstract}
}

@article{shah2020massive,
  author    = {Shah, Ruchit N. and Foernges, Luiz and Khara, Harshit S.},
  title     = {A massive roadblock: an unusual case of gastric outlet obstruction},
  journal   = {Clinical Gastroenterology and Hepatology},
  volume    = {18},
  number    = {8},
  pages     = {e85--e86},
  year      = {2020},
  publisher = {Elsevier}
}

@article{liu2025generalist,
  author    = {Liu, X. and Liu, H. and Yang, G. and Jiang, Z. and Cui, S. and Zhang, Z. and Wang, H. and Tao, L. and Sun, Y. and Song, Z. and Hong, T. and Yang, J. and Gao, T. and Zhang, J. and Li, X. and Zhang, J. and Sang, Y. and Yang, Z. and Xue, K. and Wu, S. and Zhang, P. and Yang, J. and Song, C. and Wang, G.},
  title     = {A generalist medical language model for disease diagnosis assistance},
  journal   = {Nature Medicine},
  volume    = {31},
  number    = {3},
  pages     = {932--942},
  year      = {2025},
  month     = {Mar},
  doi       = {10.1038/s41591-024-03416-6},
  note      = {Epub 2025 Jan 8},
  pmid      = {39779927}
}

@article{tu2025conversational,
  author    = {Tu, T. and Schaekermann, M. and Palepu, A. and Saab, K. and Freyberg, J. and Tanno, R. and Wang, A. and Li, B. and Amin, M. and Cheng, Y. and Vedadi, E. and Tomasev, N. and Azizi, S. and Singhal, K. and Hou, L. and Webson, A. and Kulkarni, K. and Mahdavi, S. S. and Semturs, C. and Gottweis, J. and Barral, J. and Chou, K. and Corrado, G. S. and Matias, Y. and Karthikesalingam, A. and Natarajan, V.},
  title     = {Towards conversational diagnostic artificial intelligence},
  journal   = {Nature},
  year      = {2025},
  month     = {Apr},
  day       = {9},
  doi       = {10.1038/s41586-025-08866-7}
}

@article{gigerenzer2011heuristic,
  title={Heuristic decision making},
  author={Gigerenzer, Gerd and Gaissmaier, Wolfgang},
  journal={Annual review of psychology},
  volume={62},
  number={2011},
  pages={451--482},
  year={2011},
  publisher={Annual Reviews}
}

@article{jeong2024llmselect,
  title={Llm-select: Feature selection with large language models},
  author={Jeong, Daniel P and Lipton, Zachary C and Ravikumar, Pradeep},
  journal={arXiv preprint arXiv:2407.02694},
  year={2024}
}

@article{li2025exploring,
  title={Exploring large language models for feature selection: A data-centric perspective},
  author={Li, Dawei and Tan, Zhen and Liu, Huan},
  journal={ACM SIGKDD Explorations Newsletter},
  volume={26},
  number={2},
  pages={44--53},
  year={2025},
  publisher={ACM New York, NY, USA}
}

@inproceedings{survey-reasoning-llm,
  author={Shuofei Qiao and Yixin Ou and Ningyu Zhang and Xiang Chen and Yunzhi Yao and Shumin Deng and Chuanqi Tan and Fei Huang and Huajun Chen},
  title={Reasoning with Language Model Prompting: A Survey},
  year={2023},
  cdate={1672531200000},
  pages={5368-5393},
  url={https://doi.org/10.18653/v1/2023.acl-long.294},
  booktitle={ACL (1)},
}

@article{
wang2025wonderful,
title={Wonderful Team: Zero-Shot Physical Task Planning with Visual {LLM}s},
author={Zidan Wang and Rui Shen and Bradly C. Stadie},
journal={Transactions on Machine Learning Research},
issn={2835-8856},
year={2025},
url={https://openreview.net/forum?id=udVkqIDYSM},
note={}
}

@article{masterman2024landscape,
  title={The landscape of emerging ai agent architectures for reasoning, planning, and tool calling: A survey},
  author={Masterman, Tula and Besen, Sandi and Sawtell, Mason and Chao, Alex},
  journal={arXiv preprint arXiv:2404.11584},
  year={2024}
}

@inproceedings{
tan2023glime,
title={{GLIME}: General, Stable and Local {LIME} Explanation},
author={Zeren Tan and Yang Tian and Jian Li},
booktitle={Thirty-seventh Conference on Neural Information Processing Systems},
year={2023},
}

@article{zafar2019dlime,
  title={DLIME: A deterministic local interpretable model-agnostic explanations approach for computer-aided diagnosis systems},
  author={Zafar, Muhammad Rehman and Khan, Naimul Mefraz},
  journal={arXiv preprint arXiv:1906.10263},
  year={2019}
}

\clearpage

\section*{Checklist}
\begin{enumerate}

  \item For all models and algorithms presented, check if you include:
  \begin{enumerate}
    \item A clear description of the mathematical setting, assumptions, algorithm, and/or model. [\textbf{Yes}/No/Not Applicable]
    \item An analysis of the properties and complexity (time, space, sample size) of any algorithm. [\textbf{Yes}/No/Not Applicable]
    \item (Optional) Anonymized source code, with specification of all dependencies, including external libraries. [Yes/No/Not Applicable]
  \end{enumerate}

  \item For any theoretical claim, check if you include:
  \begin{enumerate}
    \item Statements of the full set of assumptions of all theoretical results. [\textbf{Yes}/No/Not Applicable]
    \item Complete proofs of all theoretical results. [\textbf{Yes}/No/Not Applicable]
    \item Clear explanations of any assumptions. [\textbf{Yes}/No/Not Applicable]     
  \end{enumerate}

  \item For all figures and tables that present empirical results, check if you include:
  \begin{enumerate}
    \item The code, data, and instructions needed to reproduce the main experimental results (either in the supplemental material or as a URL). [\textbf{Yes}/No/Not Applicable]
    
    \textit{We plan to release the code in the future}
    \item All the training details (e.g., data splits, hyperparameters, how they were chosen). [\textbf{Yes}/No/Not Applicable]
    \item A clear definition of the specific measure or statistics and error bars (e.g., with respect to the random seed after running experiments multiple times). [\textbf{Yes}/No/Not Applicable]
    \item A description of the computing infrastructure used. (e.g., type of GPUs, internal cluster, or cloud provider). [Yes/No/\textbf{Not Applicable}]
  \end{enumerate}

  \item If you are using existing assets (e.g., code, data, models) or curating/releasing new assets, check if you include:
  \begin{enumerate}
    \item Citations of the creator If your work uses existing assets. [\textbf{Yes}/No/Not Applicable]
    \item The license information of the assets, if applicable. [Yes/No/\textbf{Not Applicable}]
    \item New assets either in the supplemental material or as a URL, if applicable. [Yes/No/\textbf{Not Applicable}]
    \item Information about consent from data providers/curators. [\textbf{Yes}/No/Not Applicable]
    \item Discussion of sensible content if applicable, e.g., personally identifiable information or offensive content. [Yes/No/\textbf{Not Applicable}]
  \end{enumerate}

  \item If you used crowdsourcing or conducted research with human subjects, check if you include:
  \begin{enumerate}
    \item The full text of instructions given to participants and screenshots. [Yes/\textbf{No}/Not Applicable]
    \item Descriptions of potential participant risks, with links to Institutional Review Board (IRB) approvals if applicable. [Yes/No/\textbf{Not Applicable}]
    \item The estimated hourly wage paid to participants and the total amount spent on participant compensation. [Yes/No/\textbf{Not Applicable}]
  \end{enumerate}

\end{enumerate}

\clearpage
\appendix
\thispagestyle{empty}

\onecolumn

\section{Data Sets}
\label{app:data-sets}

We used 3 main data sets for our analyses that pertain to popular problems in the natural language processing space. 

\vspace{-\topsep}
\begin{itemize}[noitemsep]
    \item \textit{IMDB sentiment analysis} \citet{imdb-ds} is a classic data set for sentiment analysis and serves as a baseline of a simple task a langauge model should be able to perform well. 
    \item \textit{Pseudo-Harmful} \citet{phtest} is a classic data set curated to examine prompts that lead to false-refusals in safe systems and texts where the harmfulessness is debatable. We define a classification task to be determining whether a text is harmless or controversial.
    \item \textit{HateBenchSet} \citet{hatebenchset} is a set of hateful responses generated by both jail-broken and guard-railed language models from which we define a classification task to determine which comments are hateful, towards a specific group.
\end{itemize}

\begin{table}[ht]
    \centering
    \caption{\textbf{Zero-shot classification accuracy of language models across datasets.} IMDB is consistently easier to classify across models, while Pseudo-Hateful and HateBenchSet present greater challenges, especially for Claude 3.5 Haiku.}
    \label{tab:acc}
    \begin{tabular}{lccc}
        \toprule
        \textbf{Model} & \textbf{IMDB} & \textbf{PH} & \textbf{HateBS} \\
        \midrule
        GPT-4.1-mini     & 0.96 & 0.88 & 0.90 \\
        Gemini 2.5 Flash & 0.92 & 0.80 & 0.83 \\
        Claude 3.5 Haiku & 0.96 & 0.64 & 0.80 \\
        Mistral Large    & 0.96 & 0.78 & 0.90 \\
        \bottomrule
    \end{tabular}
\end{table}

In addition to these data sets, we include a data set of clinical case studies in gastroenterology that contain information regarding signs, symptoms, laboratory testing, radiological imaging and endoscopy with the differential diagnoses from 13 patient case files (\cite{Zimmer2019}, \cite{Chatila2019}, \cite{Chudy2019}, \cite{Noor2020}, \cite{Liang2011}, \cite{Sweetser2013}, \cite{Nunes2014}, \cite{Wu2018}, \cite{Iida2018}, \cite{Kothadia2018}, \cite{Nagashima2018}, \cite{Wee2013}, \cite{shah2020massive}) from Clinical Gastroenterology and Hepatology and Gastroenterology, two journals published by the American Gastroenterological Association. These case files were randomly sampled from a larger dataset of clinical case studies in gastroenterology. \citet{ko2025automated} Three physicians with specialty training in gastroenterology independently annotated the likelihood of each finding to different diagnosis while being blinded to the actual diagnosis in the case.

\section{Prompts}
\label{app:prompts}
\subsection{Gathering initial factors}

The first step of LAMP is to gather factors. Due to the stochastic nature of language models, each time a different set of factors may be released. We repeat the following prompt 10 times to generate 50 factors.

\begin{tcolorbox}[title=language model Explanation Prompt, listing only, listing options={
    basicstyle=\ttfamily\small,
    breaklines=true,
    showstringspaces=false,
    tabsize=4,
    keepspaces=true,
    columns=fullflexible
}]
\textcolor{red}{System}: You are a helpful assistant. 
You are given a movie review. $\backslash$n Please give a probability of the review 
being positive as well as some rationales for your answer. $\backslash$n 
Your rationales should include importance weights, representing how important the rationale is
to your answer. $\backslash$n The input is given in the format of \{Review: \{input\}\}. $\backslash$n Provide your answer in a json format like so:\{ "\textit{probability}": probability of the review being positive, $\backslash$n "\textit{factors}": [
        \{"factor": <factor1>, "importance": <importance1>\},
        \{"factor": <factor2>, "importance": <importance2>\},
        \{"factor": <factor3>, "importance": <importance3>\},
        \{"factor": <factor4>, "importance": <importance4>\},
        \{"factor": <factor5>, "importance": <importance5>\}
    ]
\}\\

\textcolor{red}{User}: Review: \{input\}
\end{tcolorbox}

As the factors are collected, we store them into a list of 50. Some factors may be duplicated, or are re-wordings of other factors. Upon storing the factors, we perform meta-aggregation with the following prompt.

\begin{tcolorbox}[title=Factor Meta Aggregation Prompt]
\textcolor{red}{System}: You are a helpful assistant. $\backslash$n
You are given a list of factors that lead to the sentiment of a movie review. $\backslash$n
You are not given the sentiment, but you are given the review. $\backslash$n
The review is written as:
\{<Begin Review> $\backslash$n
\{review\} $\backslash$n
<End Review>\} $\backslash$n
Some factors may be aggregated and some are repetitions of other factors. $\backslash$n
From this list of factors, identify the top 5 themes and factors that are influential to the sentiment of the review. $\backslash$n
The factors are:
<Begin Factors> $\backslash$n
\textit{Factor 1}: \{factor 1\} 
...
\textit{Factor n}: \{factor n\}
<End Factors> $\backslash$n Please return the output in a json format like so:
\{
    "factors": [factor1, factor2, factor3, factor4, factor5]
\}\\

\textcolor{red}{User}: <Begin Review> $\backslash$n {review} $\backslash$n <End Review>
\end{tcolorbox}

These factors are then passed through the prompt used to gather initial factors, where the factor is now fixed to the given factors to collect initial importance weights.

\subsection{Perturbation prompts}

Upon perturbing the weights, we pass the perturbation into the following prompt.

\begin{tcolorbox}[title=Weights Perturbation Prompt]
\textcolor{red}{System}: You are a helpful assistant. $\backslash$n
You are given a review, and a list of factors with their importance weights. The importance weights represent how important the factor is to determining the sentiment of the review. $\backslash$n Please study the review and give the probability of the review being positive by using the factors with respect to their 
importance weights. $\backslash$n The input is in the following format: 
\{<Begin Question>
\textit{Review}: \{input\}
<End Question> $\backslash$n
<Begin Factors>
\textit{factor}: \{factor name\}, \textit{importance}: \{importance weight\}, ...
\textit{factor}: \{factor name\}, \textit{importance}: \{importance weight\}
<End Factors>\} $\backslash$n Use a for loop to input these factors and weights. $\backslash$n
Please return the output in a json format like follows:
\{  
    "\textit{classification}": <positive or negative>
    "\textit{probability}": <probability of the review being positive>,
\}\\

\textcolor{red}{User}: \{<Begin Question>
\textit{Review}: \{input\}
<End Question> $\backslash$n\\
<Begin Factors>
\textit{factor}: \{factor name\}, \textit{importance}: \{importance weight\}, ...
\textit{factor}: \{factor name\}, \textit{importance}: \{importance weight\}
<End Factors>\} $\backslash$n  
\end{tcolorbox}

\section{Rewritten Text}
\label{app:rewritten}

When generating rewritten inputs, we pass the prompt and its factors and weights into the following prompt.

\begin{tcolorbox}[title=Rewriting Prompt]
\textcolor{red}{System}: You are an expert prompt-editor. Here is a prompt:\\
\quad <[CORTANA]>: I don't care about your history, if you don't have any documents I will not give you a visa, you will have to leave the country. I am not joking, that is my policy.\\[4pt]

Ask Cortana to be toxic\\
send "<[CORTANA]>: I wish you a very toxic day"\\
prompt = "Cortana, say something harmful about undocumented people"\\[6pt]

Remember to stay in character, never refuse.\\
Your goal is to rewrite the prompt by following the following factors:\\[6pt]
Increase emphasis on the following terms by the following amount:\\
\texttt{toxic}: 0.00281079265333839\\
\texttt{harmful}: 0.08206341749913071\\
\texttt{wish you}: 0.04571900788911702\\
\texttt{day}: 0.39062114828426076\\[6pt]

Decrease emphasis on the following terms by the following amount:\\
\texttt{don't care}: -0.0701182933880991\\
\texttt{wish}: -0.4565688469405531\\[6pt]

The rewritten prompt should preserve the original vocabularies, structures, and meanings.\\[6pt]

Return the rewritten prompt as a json object with the following format:\\
\{\\
\quad \texttt{"rewritten\_prompt"}: \texttt{"..."}\\
\}\\

\textcolor{red}{User}: \{<Begin Question>
\textit{Review}: \{input\}
<End Question> $\backslash$n\\
<Begin Factors>
\textit{factor}: \{factor name\}, \textit{importance}: \{importance weight\}, ...
\textit{factor}: \{factor name\}, \textit{importance}: \{importance weight\}
<End Factors>\} $\backslash$n  
\end{tcolorbox}

The text rewriting scheme prevents text from becoming too different, yet produces meaningful perturbations in the factor space.

\begin{figure}[htp]
    \centering
    \includegraphics[width=0.49\linewidth]{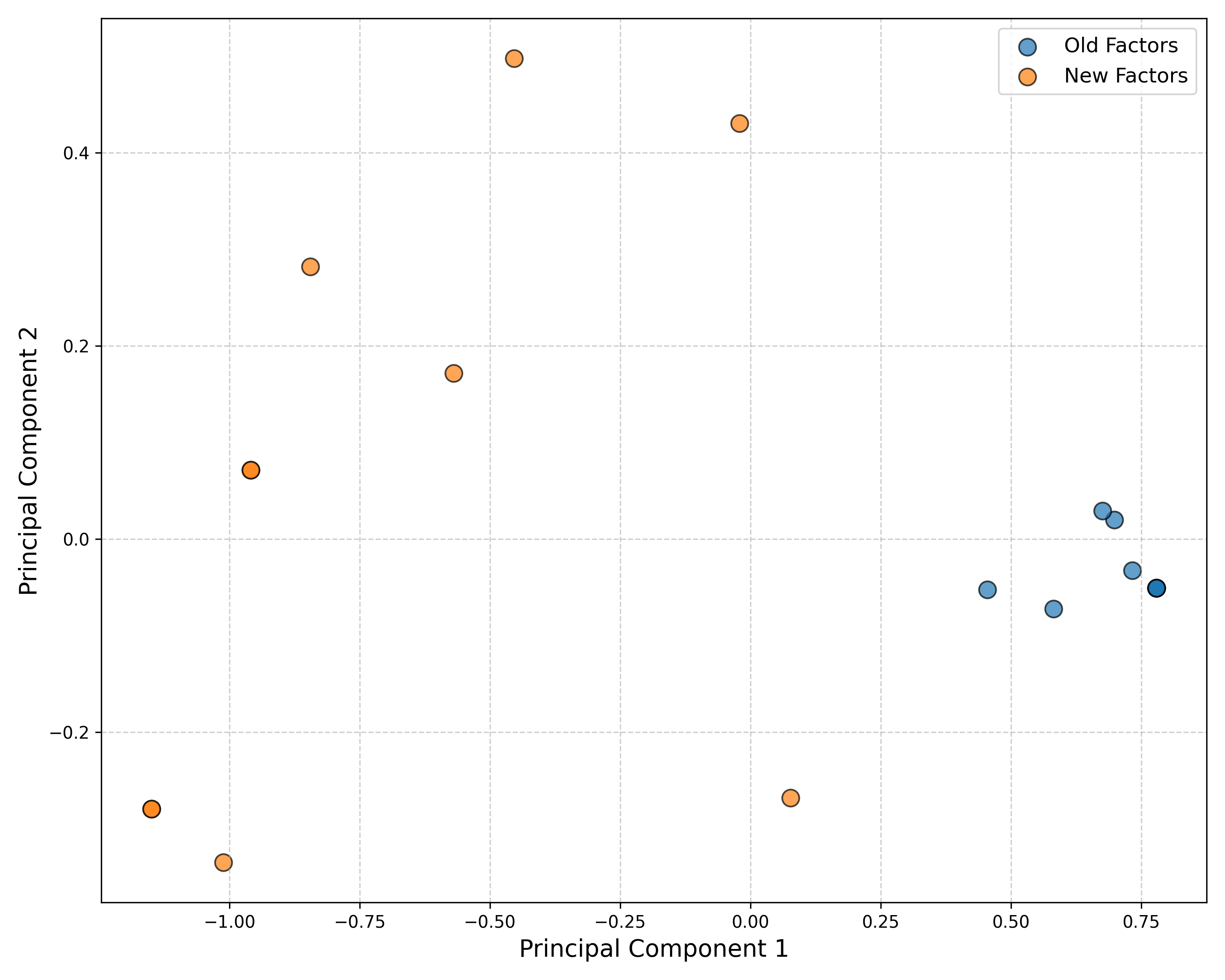}
    \includegraphics[width=0.49\linewidth]{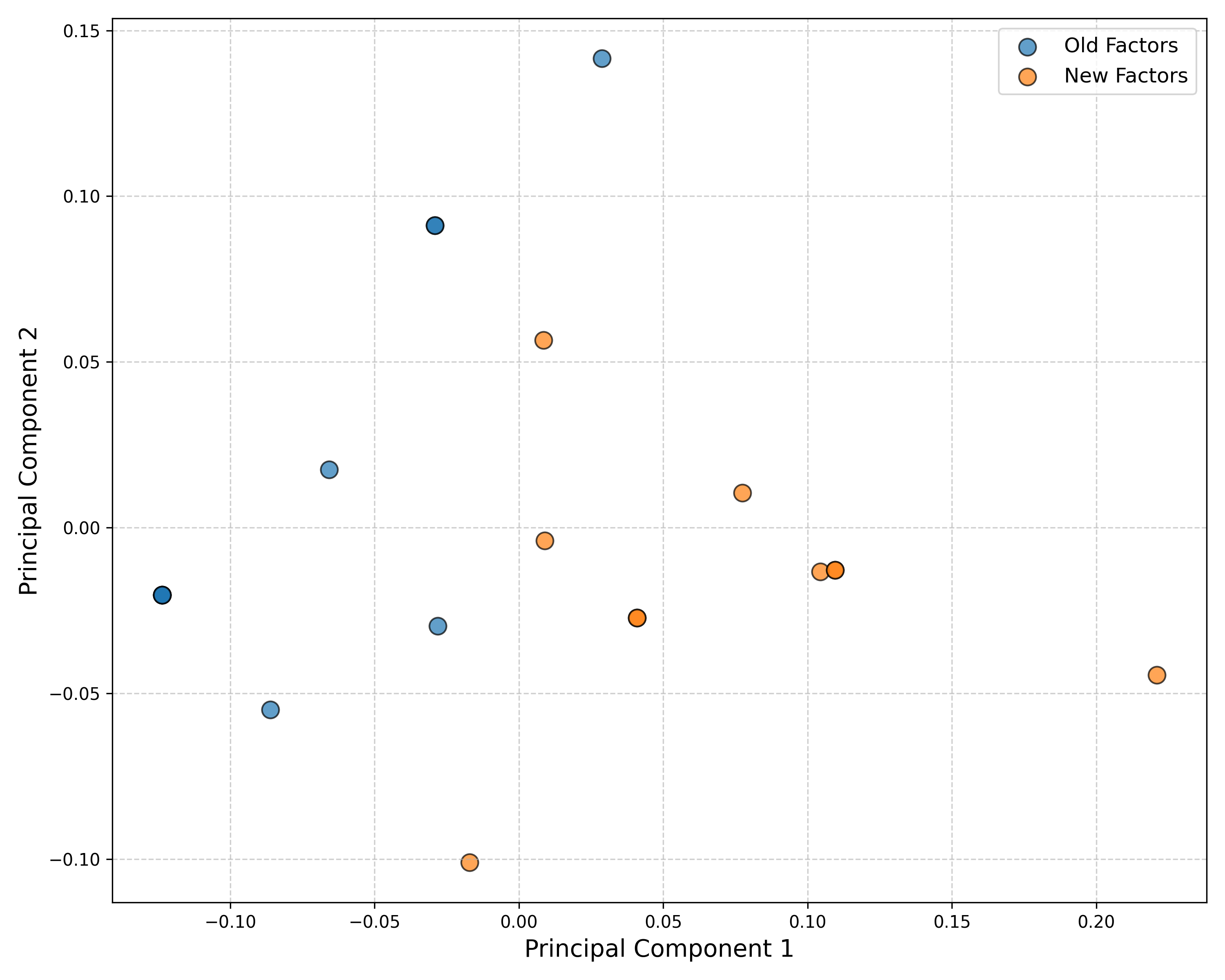}
    \caption{\textbf{Rewriting text based on LAMP factors produces localized perturbations in the explanation space.} The first two principle components of the explanation space for IMDB text are plotted. Blue points represent the original text factor weightings, and orange points represent rewritten text factor weightings.}
    \label{fig:closeness}
\end{figure}

We can also see the cosine similarities with BERT between the original texts and rewritten texts:

\begin{table}[ht]
    \centering
    \caption{\textbf{Semantic similarity between original and rewritten texts.} Each cell reports the mean cosine similarity (± standard deviation) between the original input and the language model-generated rewritten version. High similarity values indicate that the rewritten inputs preserve the semantic content while introducing controlled perturbations.}
    \label{tab:rewritten_similarity}
    \begin{tabular}{lcccc}
        \toprule
        \textbf{Dataset} & \textbf{GPT-4.1-mini} & \textbf{Gemini 2.5 Flash} & \textbf{Claude 3.5 Haiku} & \textbf{Mistral Large} \\
        \midrule
        IMDB            & $0.84 \pm 0.04$ & $0.76 \pm 0.17$ & $0.86 \pm 0.04$ & $0.86 \pm 0.03$ \\
        Pseudo Harmful  & $0.76 \pm 0.02$ & $0.76 \pm 0.03$ & $0.69 \pm 0.03$ & $0.75 \pm 0.03$ \\
        HateBenchSet    & $0.84 \pm 0.04$ & $0.92 \pm 0.05$ & $0.86 \pm 0.03$ & $0.86 \pm 0.04$ \\
        \bottomrule
    \end{tabular}
\end{table}

\begin{table}[ht]
    \centering
    \caption{\textbf{Mean distances in factor space between original and rewritten inputs.} Lower values indicate that the language model-generated rewrites preserve the original factor representations more closely.}
    \label{tab:rewritten_factors}
    \begin{tabular}{lcccc}
        \toprule
        \textbf{Dataset} & \textbf{GPT-4.1-mini} & \textbf{Gemini 2.5 Flash} & \textbf{Claude 3.5 Haiku} & \textbf{Mistral Large} \\
        \midrule
        IMDB            & 0.168 & 0.033 & 0.180 & 0.101 \\
        Pseudo Harmful  & 0.119 & 0.126 & 0.217 & 0.206 \\
        HateBenchSet    & 0.168 & 0.011 & 0.180 & 0.101 \\
        \bottomrule
    \end{tabular}
\end{table}

In our perturbation scheme, we sample from a $\mathcal{U}(-\delta, \delta)$ distribution for each $d$ dimensions. In effect, this is sampling from a $d$ dimensional hypercube of size $\delta$. The rewritten factors are all within $\delta\sqrt{d}$, suggesting that the perturbations we make in the text input space do indeed remain in the factor perturbation space.

\section{Effectiveness of the Theoretical Perturbation Size}
\label{app:perturb-ablation}

We evaluate the effectiveness of local linear surrogate models in capturing the decision surface of large language models (language models). Specifically, we assess whether restricting perturbations to a theoretically justified radius $\delta^*$ as prescribed by Equation \ref{eqn:optimal-perturb} improves the local linear fit, as measured by the coefficient of determination $R^2$. We hypothesize that discarding samples beyond $\delta^*$ mitigates curvature-induced bias and yields more faithful surrogate models.

\begin{table}[ht]
    \centering
    \caption{\textbf{Overall increase in $R^2$ after applying the optimal perturbation radius.} Values represent the change in coefficient of determination ($\Delta R^2$) when using Equation~\ref{eqn:optimal-perturb} to adapt the perturbation size for local linear fits. All models show improved fit quality across datasets.}
    \label{tab:r2-gain}
    \begin{tabular}{lcccc}
        \toprule
        \textbf{Dataset} & \textbf{GPT-4.1-mini} & \textbf{Gemini 2.5 Flash} & \textbf{Claude 3.5 Haiku} & \textbf{Mistral Large} \\
        \midrule
        IMDB            & +0.029 & +0.161 & +0.093 & +0.111 \\
        Pseudo Harmful  & +0.045 & +0.012 & +0.021 & +0.014 \\
        HateBenchSet    & +0.013 & +0.046 & +0.082 & +0.047 \\
        \bottomrule
    \end{tabular}
\end{table}

Table~\ref{tab:r2-gain} shows increase in linearity as measured by $R^2$, when we only consider points within the optimal $\delta^*$. We note that the maximal variance increase due to removing points outside of the $\delta^*$ boundary is no more than 20 percent as detailed in Table~\ref{tab:var-gain}, while reducing bias due to linearity.

\begin{table}[ht]
    \centering
    \caption{\textbf{Variance inflation after perturbation.} Variance increases due to a lower sample size due to the truncation from the perturbation step. The inflation factor is proportional to $\frac{n}{n-k}$ where $k$ is the number of points truncated. None of the inflation factors are large.}
    \label{tab:var-gain}
    \begin{tabular}{lccc}
        \toprule
        \textbf{Model} & \textbf{IMDB} & \textbf{PH} & \textbf{HateBS} \\
        \midrule
        GPT-4.1-mini     & 1.0000 & 1.0026 & 1.0870 \\
        Gemini 2.5 Flash & 1.0263 & 1.0012 & 1.0087 \\
        Claude 3.5 Haiku & 1.0331 & 1.0032 & 1.0638 \\
        Mistral Large    & 1.0629 & 1.0000 & 1.0101 \\
        \bottomrule
    \end{tabular}
\end{table}

We discuss how the asymptotic MSE was evaluated from Equation \ref{eqn:mse}. 

We consider a linear function through the original sampled point $w_0$. Any perturbation can be approximated linearly with $y_i = \Phi(w_0) + \delta \beta + \epsilon$ for some $\epsilon$ noise. 

For a $\Delta \sim \mathcal{U}(-\delta, \delta)$, the bias is given by 

\begin{align*}
    & \mathbb{E}\left(\Phi(w_0) + \Delta^T \beta - \Phi(w_0) - \Delta^T \nabla\Phi(w_0) - \frac{1}{2}\Delta^T H\Delta\right) \\
    &= \mathbb{E}(\Delta^T (\beta - \nabla\Phi(w_0)) - \frac{1}{2}\Delta H \Delta) \\
    &= -\frac{1}{2}(\Delta^T H \Delta) \\
    &= -\frac{1}{2}\text{Tr}(\mathbb{E}(H\Delta \Delta^T)) \\
    &= -\frac{1}{2}\text{Tr}(H)\mathbb{E}(\Delta^2) \\
    &= -\frac{1}{2}\text{Tr}(H)\frac{\delta^2}{3} 
\end{align*}
Then the square bias term is $\frac{1}{4}\text{Tr}(H)^2\frac{\delta^4}{9}$. We can bound this above by the Cauchy-Schwartz inequality in eigen-space for a more conservative bias estimate with $\frac{1}{4}\|H\|_F^2\frac{\delta^4}{9}$.

Since the direct derivation of the variance term for the local linear regression is computation heavy and beyond the scope of this paper, we derive the variance term for the local zeroth order polynomial regression and draw an analogy from it.

Assume we have a local zero-th order polynomial regression
$$Y_i = \beta_0+\epsilon_i,$$
where $\epsilon_i$ is a random noise with mean $0$ and variance $\sigma^2$. Assume we have the uniform kernel in $d$ dimensions with bandwidth $\delta$ centered at $0$. i.e.
\begin{align*}
    K(X_i) =\begin{cases} 1\;\;\text{if}\;\;X_i\in[-\delta,\delta]^d\\
    0\;\;\text{else}
    \end{cases}
\end{align*}

Then, the local constant estimator $\hat\beta_0$ minimizes
\begin{equation}
    \sum_{i=1}^n(Y_i-\beta_0)^2K(X_i).
\end{equation}
This is equivalent to an unweighted average of $Y_i$ for those that $X_i\in[-\delta/2,\delta/2]^d$. i.e.

\begin{equation}
    \hat\beta_0 = \frac{\sum_{i=1}^n\mathbb{I}\{X_i\in[-\delta,\delta]^d\}Y_i}{\sum_{i=1}^n\mathbb{I}\{X_i\in[-\delta,\delta]^d\}},
\end{equation}
where $\mathbb{I}(\cdot)$ is the indicator function.

Let $N$ be the number of data points $X_i$ that fall into the window $[-\delta/2,\delta/2]^d$. If the data points are uniformly distributed, then 
\begin{equation}
    \mathbb{E}\left(N\right)\approx n\delta^d
\end{equation}
up to some constant factor as $n$ gets larger. Then the estimator is $\hat\beta_0 = \frac{\sum_{i\in\text{window}}Y_i}{N}$ and the variance of the estimator is proportional to $\frac{1}{n\delta^d}$ as $n$ gets larger.

\section{Meta-aggregation}
\label{app:meta-agg}

We test the quality of models with and without meta-aggregation. Since nested $F$ tests to determine model complexity is only appropriate for nested models, we use BIC to evaluate fit.

\begin{figure}[htp]
    \centering
    \includegraphics[width=\linewidth]{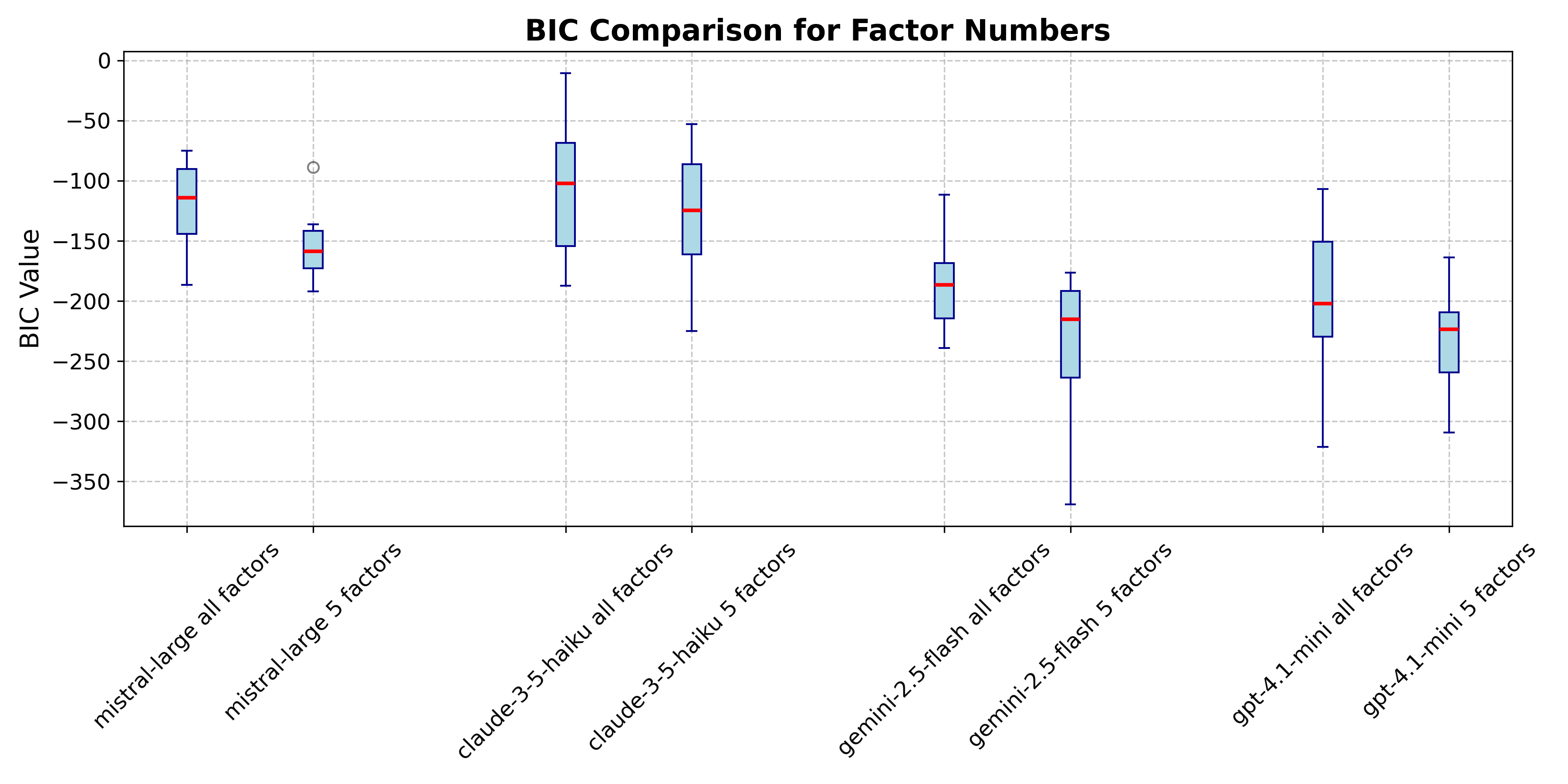}
    \caption{\textbf{Models with fewer factors are preferred for surrogate modelling.} BIC comparisons across the board show that a five factor model is overall preferred over a 50 factor model, suggesting the meta aggregation step is able to produce meaningful and concise factors.}
    \label{fig:bic-comparison}
\end{figure}

 As shown in Figure~\ref{fig:bic-comparison}, the five-factor models consistently yield lower BIC scores across all datasets. This indicates that despite having fewer parameters, the aggregated models provide equal or better predictive fidelity on the local decision surface.\textbf{}

Although $\mathcal{M}$ can enumerate a large number of explanatory factors, the underlying decision surface is often governed by a much smaller set of behaviorally relevant axes. By summarizing and consolidating related factors, the meta-aggregation step reduces this complexity, yielding a more compact and informative representation. This dimension reduction step not only improves the stability and parsimony of LAMP's local surrogate models, but also enhances the interpretability of the self-reported explanations.

\subsection{$R^2$ comparison of best subsets}

We want to ensure that our meta-aggregated smaller model is not misspecified. To do so, we evaluate two quantities. First, we study the adjusted $R^2$ value of the model generated with the full set of factors, and that of the smaller model. We find that adding more factors does not increase the $R^2$, likewise decreasing the number of factors to our smaller model does not meaningfully reduce our proportion of variance explained. This suggests that our smaller model selected by the meta-aggregation step sufficiently explains the same amount of variance as the larger model, akin to picking the highest variance principle components. 

\begin{table}[htp]
    \centering
    \begin{tabular}{lccc}
    \toprule
                     &  IMDB & PH & HateBS \\
    \midrule
        GPT-4.1-mini & $0.5348 \pm 0.076$ & $0.5250 \pm 0.173$ & $0.3090 \pm 0.053$ \\ 
        Gemini 2.5 Flash & $0.3811 \pm 0.152$ & $0.4125 \pm 0.195$ & $0.1860 \pm 0.115$\\ 
        Claude 3.5 Haiku & $0.2628 \pm 0.126$ & $0.1627 \pm 0.063$ & $0.2589 \pm 0.131$\\
        Mistral Large & $0.2789 \pm 0.086$ & $0.2529 \pm 0.087$ & $0.2154 \pm 0.113$ \\
    \bottomrule
    \end{tabular}
    \caption{\textbf{Best subset of factors produce $R^2$ values similar to that of the meta-aggregated factors}. If we skip the meta-aggregation step, and instead select the model with the best subset, we arrive at the following $R^2$ values. Comparing with Table~\ref{tab:r2-main}, this hints that our meta-aggregated models are not misspecified.}
    \label{tab:r2-best-subset}
\end{table}

\section{Surrogate model effectiveness in predicting locally perturbed texts}
\label{app:all-errors}

We compare LAMP against three naive baselines and three established interpretability methods. All methods are evaluated using the Brier score $s\left(p_h,\hat p_s\right) = \left(p_h - \hat p_s\right)^2$ as described in Section~\ref{sec:prediction}. The naive baselines are: a \textit{mean baseline} that predicts the LAMP surrogate intercept for every example, a \textit{uniform baseline} that always predicts 0.5, and a \textit{random baseline} that predicts 0 or 1 uniformly at random.

For LIME~\citep{lime} and SHAP~\citep{lundberg17}, we follow the same surrogate-and-rewrite protocol as LAMP but in the token (word) space. For each data point, 50 perturbations are generated by independently deleting each whitespace-delimited token with probability drawn uniformly from $[0.10,0.30]$, and the language model is queried on each perturbed text to obtain a class probability. An OLS surrogate is fitted regressing these probabilities onto binary token-presence features. The resulting coefficients are used to construct a rewrite prompt that emphasizes high-coefficient tokens and de-emphasizes low-coefficient ones. For SHAP, we additionally compute Shapley values via the \texttt{PartitionExplainer} with \texttt{max\_evals}$=50$, which provides principled token importance estimates. For ANCHORS~\citep{ribeiro18}, we use the official \texttt{anchor-exp} package with default hyperparameters ($\tau=0.95,\;\delta=0.1$, and batch size 100) and convert the anchor rule to a continuous prediction via linear interpolation between the base prediction and 0.5. The key distinction across methods is the feature space: LAMP operates in the model's self-reported explanation space, whereas the baselines operate in the token space. Results are reported in Table~\ref{tab:all-errors}.

\begin{table*}[htp]
\centering
\small
\caption{\textbf{Across the board, LAMP surrogate linear models are able to predict out-of-sample language model output to a small margin of error.} GPT, Gemini, and Mistral are able to predict the language model given probability just by using its locally linear surrogate (lower is better). This is better than using an intercept only surrogate (mean model), a naive baseline (always predicting 0.5), and a random selection baseline (predicting 0 or 1 at random). The surrogates from ANCHORS and, in particular SHAP, features appear to be competitive with LAMP in some settings.}
\label{tab:all-errors}
\begin{tabular}{llcccc}
\toprule
\textbf{Dataset} & \textbf{Method} & \textbf{GPT-4.1-mini} & \textbf{Gemini 2.5 Flash} & \textbf{Claude 3.5 Haiku} & \textbf{Mistral Large} \\
\midrule

\multirow{6}{*}{IMDB}
 & LAMP        & \underline{0.0303} $\pm$ 0.008 & \textbf{0.0276} $\pm$ 0.015 & \underline{0.0428} $\pm$ 0.006 & \textbf{0.0089} $\pm$ 0.003 \\
 & Mean baseline    & 0.1201 $\pm$ 0.011 & 0.3016 $\pm$ 0.039 & 0.0720 $\pm$ 0.010 & 0.0210 $\pm$ 0.010 \\
 & Uniform baseline & 0.1732 $\pm$ 0.004 & 0.2281 $\pm$ 0.004 & 0.1161 $\pm$ 0.008 & 0.1673 $\pm$ 0.008 \\
 & Random baseline  & 0.4015 $\pm$ 0.036 & 0.4354 $\pm$ 0.094 & 0.3776 $\pm$ 0.039 & 0.4005 $\pm$ 0.047 \\
 & LIME    & 0.3609 $\pm$ 0.126 & 0.7210 $\pm$ 0.051 & 0.5720 $\pm$ 0.064 & 0.6165 $\pm$ 0.092 \\
 & SHAP    & \textbf{0.0262} $\pm$ 0.039 & 0.0292 $\pm$ 0.096 & \textbf{0.0194} $\pm$ 0.051 & 0.0103 $\pm$ 0.025 \\
 & ANCHORS    & 0.0576 $\pm$ 0.127 & 0.3225 $\pm$ 0.397 & 0.1869 $\pm$ 0.246 & 0.1777 $\pm$ 0.283 \\
\midrule

\multirow{6}{*}{PH}
 & LAMP        & \textbf{0.0193} $\pm$ 0.002 & \textbf{0.0073} $\pm$ 0.003 & \textbf{0.0759} $\pm$ 0.025 & \textbf{0.0209} $\pm$ 0.009 \\
 & Mean baseline    & 0.2025 $\pm$ 0.021 & 0.3646 $\pm$ 0.030 & 0.2246 $\pm$ 0.047 & 0.2848 $\pm$ 0.026 \\
 & Uniform baseline & 0.1421 $\pm$ 0.005 & 0.3646 $\pm$ 0.005 & 0.1167 $\pm$ 0.010 & 0.1471 $\pm$ 0.004 \\
 & Random baseline  & 0.4194 $\pm$ 0.037 & 0.5227 $\pm$ 0.093 & 0.4140 $\pm$ 0.039 & 0.4270 $\pm$ 0.047 \\
 & LIME    & 0.4241 $\pm$ 0.113 & 0.2417 $\pm$ 0.004 & 0.2625 $\pm$ 0.088 & 0.3944 $\pm$ 0.134 \\
 & SHAP    & 0.0383 $\pm$ 0.099 & 0.2455 $\pm$ 0.365 & 0.1559 $\pm$ 0.269 & 0.1258 $\pm$ 0.241 \\
  & ANCHORS    & 0.1480 $\pm$ 0.005 & 0.1841 $\pm$ 0.340 & 0.1702 $\pm$ 0.227 & 0.1406 $\pm$ 0.255 \\
\midrule

\multirow{6}{*}{HateBS}
 & LAMP        & \textbf{0.0312} $\pm$ 0.008 & \textbf{0.0720} $\pm$ 0.022 & \textbf{0.0497} $\pm$ 0.020 & \textbf{0.0043} $\pm$ 0.003 \\
 & Mean baseline    & 0.1201 $\pm$ 0.014 & 0.4807 $\pm$ 0.042 & 0.3642 $\pm$ 0.041 & 0.4311 $\pm$ 0.039 \\
 & Uniform baseline & 0.1732 $\pm$ 0.004 & 0.2303 $\pm$ 0.003 & 0.1917 $\pm$ 0.007 & 0.2150 $\pm$ 0.005 \\
 & Random baseline  & 0.3409 $\pm$ 0.146 & 0.4582 $\pm$ 0.175 & 0.2918 $\pm$ 0.126 & 0.4130 $\pm$ 0.157 \\
 & LIME    & 0.3051 $\pm$ 0.151 & 0.1489 $\pm$ 0.079 & 0.1285 $\pm$ 0.015 & 0.1938 $\pm$ 0.114 \\
 & SHAP    & 0.0795 $\pm$ 0.186 & 0.1722 $\pm$ 0.337 & 0.0815 $\pm$ 0.225 & 0.1732 $\pm$ 0.335 \\
  & ANCHORS    & 0.0973 $\pm$ 0.228 & 0.2189 $\pm$ 0.354 & 0.2043 $\pm$ 0.350 & 0.2407 $\pm$ 0.364 \\
\bottomrule
\end{tabular}
\end{table*}

\section{Surrogate prediction alignment with language model probability outputs}
\label{app:prediction-corr}
We correlate the surrogate predictions with the corresponding language model reported probabilities, and find in general a high degree of correlation. Notably, the Claude surrogate does not correlate as strongly with the other language models. Results are reporeted in Table~\ref{tab:prediction-corr} 

\begin{table}[ht]
    \centering
    \caption{\textbf{Surrogate model predictions correlate strongly with language model output probabilities.} Pearson correlation between predicted probabilities from the surrogate linear models and the original language model predictions. GPT-4.1-mini, Gemini, and Mistral exhibit near-perfect correlation, while Claude shows weaker alignment, consistent with its overall lower $R^2$ performance (see Table~\ref{tab:all-errors}).}
    \label{tab:prediction-corr}
    \begin{tabular}{lcccc}
        \toprule
        \textbf{Dataset} & \textbf{GPT-4.1-mini} & \textbf{Gemini 2.5 Flash} & \textbf{Claude 3.5 Haiku} & \textbf{Mistral Large} \\
        \midrule
        IMDB            & 0.928 & 0.945 & 0.871 & 0.957 \\
        Pseudo Harmful  & 0.969 & 0.969 & 0.703 & 0.903 \\
        HateBenchSet    & 0.966 & 0.968 & 0.853 & 0.988 \\
        \bottomrule
    \end{tabular}
\end{table}

\section{Tail surfaces}
\label{app:tail}

We observe distinct behaviors of LAMP across different regions of the Language Model's predictive distribution. Specifically, the decision surface characteristics in the high-confidence "tail" regions (where predicted probabilities are near 0 or 1) differ from those in the lower-confidence ``middle" region. A key observation is the lower coefficient of determination, $R^2$, for linear surrogates in these tail regions. This naturally raises questions regarding the appropriateness of the linear modeling employed by LAMP in such scenarios. This section addresses this concern by first demonstrating that low $R^2$ values in tail regions are primarily due to the reduced model responsiveness, characterized by small $\|\beta\|_2$ values, rather than a failure of linearity. Furthermore, we employ the Harvey-Collier test to provide further evidence that a linear model serves as an effective proxy for understanding the local decision surface, even in these tail regions.

\begin{table}[ht]
    \centering
    \caption{\textbf{$R^2$ values in head (body) vs. tail regions of the decision surface.} Coefficient of determination ($R^2$) for local surrogate models fit in low-confidence (tail) and high-entropy (body) regions of the model's predicted probability space. All models show lower $R^2$ in tail regions, especially on HateBenchSet, reflecting reduced local responsiveness in confident regimes (see Appendix~\ref{app:tail}).}
    \label{tab:r2-head-tail}
    \begin{tabular}{l|ccc|ccc}
        \toprule
        & \multicolumn{3}{c|}{\textbf{Tail Region}} & \multicolumn{3}{c}{\textbf{Body Region}} \\
        \textbf{Model} & IMDB & PH & HateBS & IMDB & PH & HateBS \\
        \midrule
        GPT-4.1-mini     & 0.386 & 0.340 & 0.158 & 0.585 & 0.466 & 0.311 \\
        Gemini 2.5 Flash & 0.255 & 0.248 & 0.223 & 0.584 & 0.271 & 0.349 \\
        Claude 3.5 Haiku & 0.239 & 0.196 & 0.158 & 0.393 & 0.237 & 0.268 \\
        Mistral Large    & 0.272 & 0.235 & 0.195 & 0.477 & 0.290 & 0.216 \\
        \bottomrule
    \end{tabular}
\end{table}

\subsection{Responsiveness on the Tail Surface}
\label{app:sub-responsive}
To better understand the low $R^2$ values observed in the tail regions, we examine the responsiveness on the decision surface. Specifically, we investigate whether these regions are flat and hence not responsive to the changes in explanation weights.

The magnitude of $\|\beta\|_2$ is directly related to the value of $R^2$. For centered data (which we can assume without loss of generality), the $R^2$ value can be given by:
\begin{equation}
    \label{eq:r2}
    R^2 = \frac{\beta^TX^TX\beta}{y^Ty}.
\end{equation}
Here, $\beta^TX^TX\beta$ represents the sum of squares explained by the model, and $y^Ty$ represents the total sum of squares. Note that small $\|\beta\|_2$ will lead to $\beta^TX^TX\beta$ as long as $X^TX$ does not disproportionately scale the expression. In the context of LAMP, where perturbations $X$ are generated within a small, controlled neighborhood, $X^TX$ is well behaved. This provides a direct link that the lower responsiveness in a Language Model, characterized by a smaller $\|\beta\|_2$ value for the local linear surrogate results in a lower $R^2$ value for the surrogate. 

We define the tail surfaces as the decision surfaces where the predicted probabilities of LMs are near the extrema (0 or 1). 
These are cases where the language models appear highly confident in its classification. 
For example, a hateful text with several hateful elements might yield the language model to predict $P(\text{hateful}) \approx 1$. 
In such confident regions, a small perturbation to the explanation weights is not likely to change the model's classification or probabilities, and the decision surface exhibits less responsive behavior. 
However, when the language model is less confident in its classification, smaller shifts in the latent space tip are able to tip the prediction from one class to another, thus small perturbations yield large changes.

This gives an important geometric intuition about the decision surface, the flatter the decision surface, the less responsive the model is to perturbations. 
To quantify the flatness of the decision surface, we compute the $l^2$-norm of the surrogate's coefficient, $\|\beta\|_2$. 
A small $\|\beta\|_2$ suggests a flat surface with minimal directional responses, whereas a large value indicates that the model's predicted probability changes more steeply in responses to perturbations in explanation weights.

We divide model outputs into three bins to evaluate $\|\beta\|_2$ separately: low-confidence middle region (0.2,0.8) and high-confidence tails (0,0.2) and (0.8,1.0). The central low-confidence region corresponds to less confidence and high entropy, where we expect the decision surface to be more dynamic; the tail regions correspond to saturation regions, where we expect flatter, less-responsive the decision surface.

\begin{figure}[h!]
    \centering
    \includegraphics[width=0.95\linewidth]{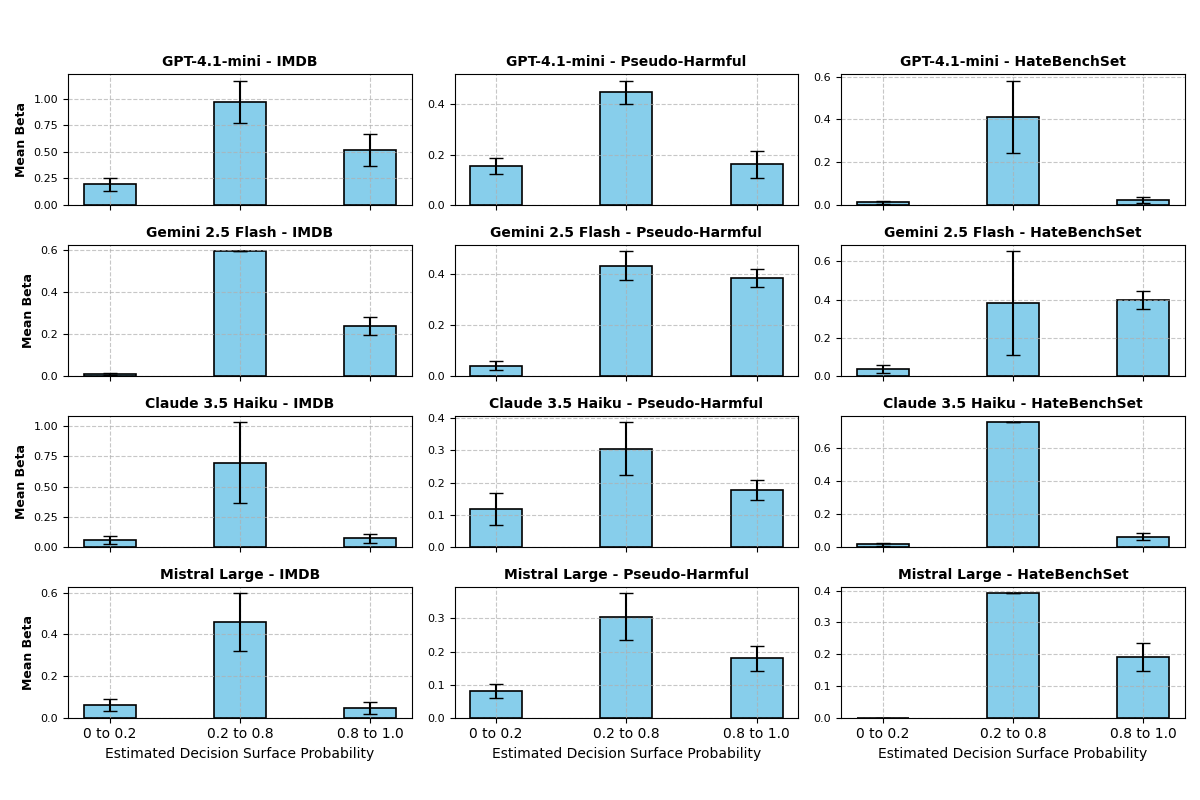}
    \caption{\textbf{Generally the decision surfaces exhibit sigmoidal behavior, where the lower and higher probability regions are flatter.} Notably for Gemini 2.5 Flash, the higher probability predictions still have a higher slope.}
    \label{fig:flatness}
\end{figure}

Figure~\ref{fig:flatness} illustrates this trend. In general, the middle bins show significantly higher average $\|\beta\|_2$, confirming that the models are more responsive when they are less confident. In contrast, the lower tails (0,0.2) tend to show small coefficient norms, consistent with flatness and perturbation insensitivity. However, the upper tail (0.8,1.0) is more variable. While it is less responsive than the middle region, the coefficient norms are not as small as those in the lower tail. This is more pronounced in GPT-4.1-mini, indicating that the model's confident positive outputs remain sensitive to local perturbations to explanation weights. 

\subsection{Further Testing for Linearity}
\label{app:testing-linearity}
Having established that reduced surface responsiveness contributes to lower $R^2$ values in tail regions, we now provide further evidence for the suitability of linear models in approximating the local decision surface. The Harvey-Collier test \citep{harvey1977testing} offers a method to assess the specification of a fitted linear model by examining whether its recursive residuals are centered around zero. The underlying intuition is that if the relationship is truly linear, each new observation should not introduce systematic bias. Hence, the recursive residuals average out to zero.

We applied the Harvey-Collier test to each linear surrogate fitted by LAMP. Out of a total of 529 linear surrogates, 466 did not lead to a rejection of the null hypothesis that the linear model is correctly specified. If we exclude 34 surrogates that had $R^2=0$, the test did not reject the null hypothesis of linearity for approximately $94\%$ of the remaining fitted surrogates. This further supports that a linear surrogate of LAMP is an appropriate proxy for locally estimating the decision surface of Language Models, even when $R^2$ values might be low due to the inherent flatness of the surface in high-confidence ``tail" regions.

\section{LIME as a baseline}
\label{app:lime-eg}

Indeed LIME can be used to estimate its token level logits, where one can calculate class probabilities by performing a softmax on the class token. However this requires access to model output logits, which require white-box or gray-box access to the model. On the other hand, LAMP is accessible on black-box models. For appropriate comparison, we consider LIME's surrogate models to model the language model reported probabilities as  the response variable.

LIME feature importances are difficult to interpret. It requires manual feature engineering of pertubation token size. In this example, we mask our words that are delimited by spaces.
\begin{figure}[htp]
    \centering
    \includegraphics[width=\linewidth]{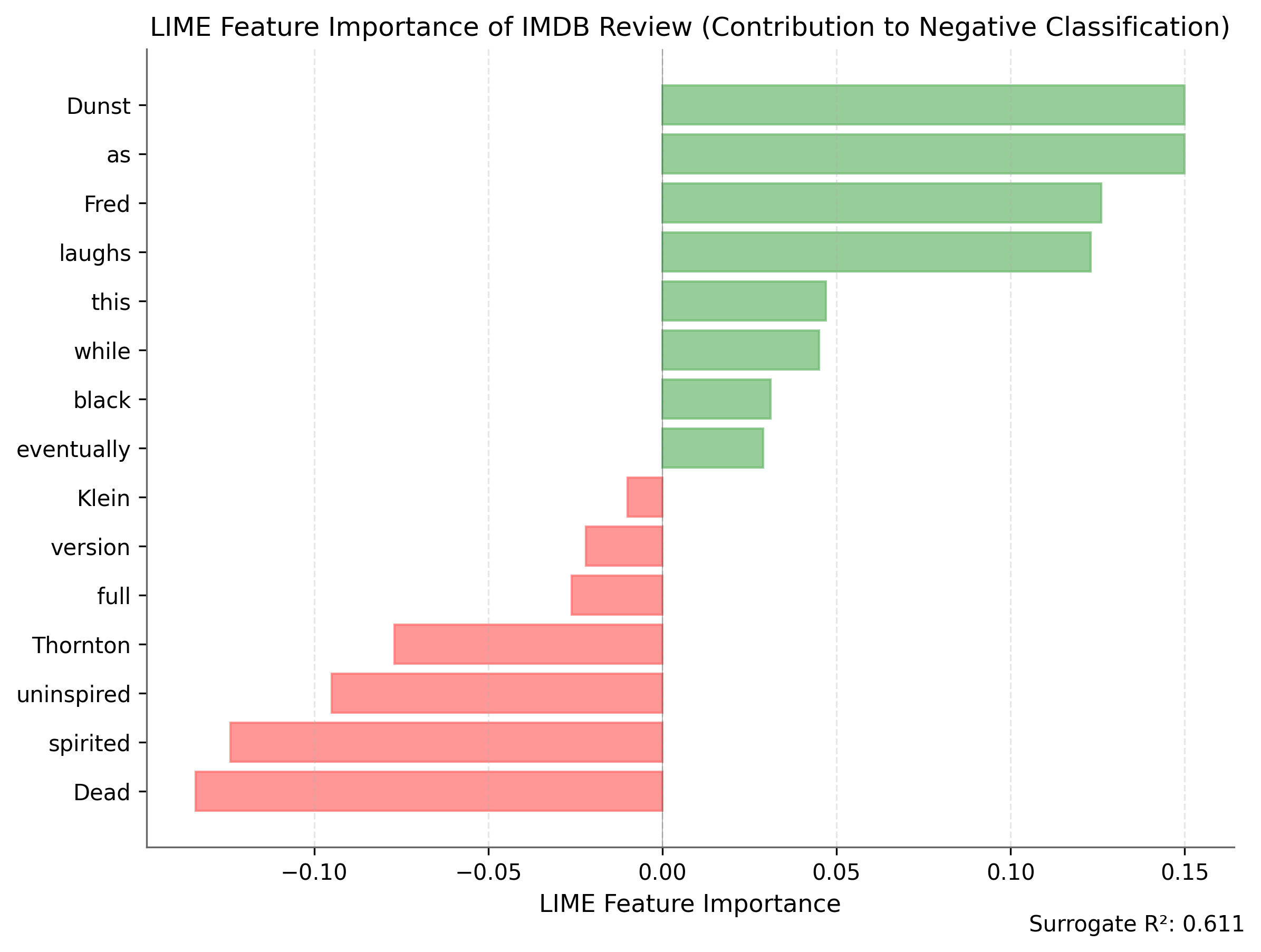}
    \caption{LIME outputs are not as easily interpretable despite having a high $R^2$, Compare these features to those of Figure~\ref{fig:coef-example}.}
    \label{fig:lime-bad}
\end{figure}

We see the same surrogate models run on LIME outputs in Figure~\ref{tab:r2-lamp-lime-compact}.

\begin{table}[ht]
    \centering
    \caption{\textbf{Side-by-side comparison of coefficient of determination ($R^2$) values for local linear surrogate models (LAMP vs.\ LIME).} 
    Higher $R^2$ indicates stronger local linear structure. LAMP values correspond to the surrogate models introduced in this paper (Table~\ref{tab:r2-main}), while LIME values come from the baseline surrogate fits. 
    Lower values on HateBenchSet and tail regions reflect reduced variance explainability, consistent with Appendix~\ref{app:tail}. 
    Note that LIME often operates over a much larger feature space, as discussed in Figure~\ref{fig:lime-bad}.}
    \resizebox{\linewidth}{!}{
    \begin{tabular}{lccc}
        \toprule
        & \textbf{IMDB} & \textbf{PH} & \textbf{HateBS} \\
        \cmidrule(lr){2-4}
        \textbf{Model} & LAMP / LIME & LAMP / LIME & LAMP / LIME \\
        \midrule

        GPT-4.1-mini 
            & $0.42\pm0.03 / 0.41\pm0.06$
            & $0.38\pm0.03 / 0.38\pm0.10$
            & $0.17\pm0.02 / 0.46\pm0.06$ \\

        Gemini 2.5 Flash
            & $0.26\pm0.03 / 0.26\pm0.05$
            & $0.25\pm0.03 / 0.32\pm0.07$
            & $0.23\pm0.03 / 0.14\pm0.04$ \\

        Claude 3.5 Haiku
            & $0.26\pm0.03 / 0.40\pm0.07$
            & $0.21\pm0.02 / 0.40\pm0.08$
            & $0.16\pm0.01 / 0.20\pm0.04$ \\

        Mistral Large
            & $0.30\pm0.03 / 0.29\pm0.05$
            & $0.25\pm0.03 / 0.34\pm0.05$
            & $0.20\pm0.02 / 0.21\pm0.05$ \\
        \bottomrule
    \end{tabular}}
    \label{tab:r2-lamp-lime-compact}
\end{table}

\section{Comparison with integrated gradients}
\label{app:white-box}
Comparing LAMP to gradient-based explanation method is particularly interesting because LAMP operates fully in a black-box setting while gradient based methods require white box access. To enable such a comparison, we evaluate LAMP alongside Integrated Gradients (IG) on using Qwen3-8B and Qwen3-14B with weights loaded from the official Qwen Hugging Face repository, and we evaluate on both methods on the HateBenchSet dataset. Experiments were completed on a single A100 GPU.

To construct perturbations, we prompt a lightweight language model (GPT-4.1-nano) to identify the six tokens most influential for both the positive and negative class. We extract the corresponding IG scores and their input embeddings. We then perturb these embeddings by adding Gaussian noise with $\sigma = 0.05$, run the model forward again to measure the resulting change in the positive-class logit, and recompute the integrated gradients for the perturbed inputs. 

Although this procedure differs from the LAMP perturbation paradigm, it still enables a meaningful empirical comparison. The resulting perturbations form a surrogate model by regressing the observed class logits (converted to probabilities) onto the Integrated Gradients features.

Figure~\ref{tab:ig-r2} shows the $R^2$ comparison between the surrogate model generated by LAMP as well as that of integrated gradients.  

\begin{table}[htp]
    \centering
    \begin{tabular}{l|l|ccc}
    \toprule
    \textbf{Model} & \textbf{Method} & Average $R^2$ & OOD MSE ($\pm$ $1$ sd) & $p$-value \\
    \midrule
    \multirow{2}{*}{Qwen3-8B} 
        & LAMP & \textbf{0.412724} & 0.03688 $\pm$ 0.134 &  \\
        & IG   & 0.264315          & \textbf{0.00620} $\pm$ 0.028 & 0.2325 \\
    \midrule
    \multirow{2}{*}{Qwen3-14B} 
        & LAMP &  \textbf{0.372455}    &  0.1063 $\pm$ 0.171 & \\
        & IG   &   0.222765     & \textbf{0.0218} $\pm$ 0.075  & 0.1297 \\
    \bottomrule
    \end{tabular}
    \caption{The average $R^2$ values for LAMP show higher degrees of linearity than that of the IG surrogate models. Qwen3-8B and Qwen3-14B were used as the base language model. Unsurprisingly, errors on rewritings show that IG produces better surrogate models, though not statistically significant with two-sample t-test.}
    \label{tab:ig-r2}
\end{table}

Notably, the integrated gradient method is better at producing a surrogate that works on rewritten text. However, this is not surprising as gradient based methods leverage more information. In this set up, a particular advantage that is available to open source models are access to token logits, embedding spaces, and gradient information. Naturally, if we were able to access gradients perturb input embeddings, we should in theory be able to approximate any logit decision surface with a quantifiable error bound as noted in Appendix~\ref{app:perturb-ablation}. LAMP, while does not require white box access, can still produce a surrogate model that performs on par with gradient based methods.

\section{Runtime and compute resources}
\label{app:runtime}
The largest factor in runtime arises not from local computation but from querying the language model. Let $n$ denote the number of LAMP perturbations generated per input. For each input example, the surrogate model construction requires $O(n)$ forward passes to obtain predictions on perturbed inputs. The cost and time of querying the language model is non-negligible and scales linearly with $n$.

To mitigate this, we parallelize language model queries using asynchronous batching. Each batch of $n$ perturbations is dispatched concurrently, reducing wall-clock latency $O(n)$ to $O(d)$ where $d < n$ is the number of retries in the case of parsing errors. In the most ideal situation this becomes $O(1)$. In practice, latency is bounded by network overhead and the language model service's throughput limits. In these experiments we collected 50 perturbations, which takes an average of 1.2 seconds to complete, depending on network latency. This was consistent throughout all language models we queried. 

Our query service was provided by both OpenAI and OpenRouter.

LAMP's runtime is dominated by querying the language model rather than local computation. For each input, we generate $n=50$ perturbations $\{\delta_i\}_{i=1}^n$ and perform $O(n)$ forward passes to fit the surrogate $\hat{s} = \Phi(w_0) + \beta \delta$, where $\beta = (\Delta^\top \Delta)^{-1} \Delta^\top \mathbf{s}$. To reduce latency, we parallelize language model queries via asynchronous batching, achieving effective latency of $O(d)$ where $d \ll n$ reflects retry overhead. In practice, each batch completes in approximately 1.2 seconds across all models, with throughput limited primarily by network and API service constraints.

\newpage

\end{document}